\documentclass[11pt]{article}
\usepackage{amsmath}
\usepackage{amssymb}
\usepackage{amscd}
\usepackage{picinpar}
\usepackage{times}
\usepackage{pb-diagram}
\usepackage{graphicx}
\usepackage{wrapfig}
\usepackage{xspace}

\newtheorem{theorem}{Theorem}[section]
\newtheorem{lemma}[theorem]{Lemma}

\newtheorem{corollary}[theorem]{Corollary}
\newtheorem{definition}[theorem]{Definition}
\newtheorem{problem}[theorem]{Problem}

\newcommand{\qed}{\nobreak \ifvmode \relax \else
      \ifdim\lastskip<1.5em \hskip-\lastskip
      \hskip1.5em plus0em minus0.5em \fi \nobreak
      \vrule height0.75em width0.5em depth0.25em\fi}

\begin{document}

\title{A Geometric View of Optimal Transportation and Generative Model}

\author{
Na Lei
\thanks{Dalian University of Technology, Dalian, China. Email: nalei@dlut.edu.cn}
\and
Kehua Su
\thanks{Wuhan University, Wuhan, China. Email: skh@whu.edu.cn}
\and
Li Cui
\thanks{Beijing Normal University, Beijing, China. Email: licui@bnu.edu.cn}
\and
Shing-Tung Yau
\thanks{Harvard University, Boston, US. Email: yau@math.harvard.edu}
\and
David Xianfeng Gu
\thanks{Stony Brook University, New York, US. Email: gu@cs.stonybrook.edu.}
}

\date{}
\maketitle


\begin{abstract}
In this work, we show the intrinsic relations between optimal transportation and convex geometry, especially the variational approach to solve Alexandrov problem: constructing a convex polytope with prescribed face normals and volumes. This leads to a geometric interpretation to generative models, and leads to a novel framework for generative models.

By using the optimal transportation view of GAN model, we show that the discriminator computes the Kantorovich potential, the generator calculates the transportation map. For a large class of transportation costs, the Kantorovich potential can give the optimal transportation map by a close-form formula. Therefore, it is sufficient to solely optimize the discriminator. This shows the adversarial competition can be avoided, and the computational architecture can be simplified.

Preliminary experimental results show the geometric method outperforms WGAN for approximating probability measures with multiple clusters in low dimensional space.

\end{abstract}

\section{Introduction}

\paragraph{GAN model}

Generative Adversarial Networks (GANs)~\cite{goodfellow2014generative} aim at learning a mapping from a simple distribution to a given distribution. A GAN model consists of a generator $G$ and a discriminator $D$, both are represented as deep networks. The generator captures the data distribution and generates samples, the discriminator estimates the probability that a sample came from the training data rather than $G$. Both generator and the discriminator are trained simultaneously. The competition drives both of them to improve their performance until the generated samples are indistinguishable from the genuine data samples. At the Nash equilibrium~\cite{zhao2016energy}, the distribution generated by $G$ equals to the real data distribution. GANs have several advantages: they can automatically generate samples, and reduce the amount of real data samples; furthermore, GANs do not need the explicit expression of the distribution of given data.

Recently, GANs receive an exploding amount of attention. For example, GANs have been widely applied to numerous computer vision tasks such as image inplainting~\cite{pathak2016context, yeh2016semantic, li2017generative}, image super resolution~\cite{ledig2016photo, iizuka2017globally}, semantic segmentation~\cite{zhu2016adversarial, luc2016semantic}, object detection~\cite{radford2015unsupervised, li2017perceptual, wang2017fast}, video prediction~\cite{mathieu2015deep, vondrick2016generating}, image translation~\cite{isola2016image, zhu2017unpaired, dong2017unsupervised, liu2017unsupervised}, 3D vision~\cite{wu2016learning, park2017transformation}, face editing~\cite{larsen2015autoencoding, liu2016coupled, perarnau2016invertible, shen2016learning, brock2016neural, shu2017neural, huang2017beyond}, etc. Also, in machine learning field, GANs have been applied to semi-supervised learning~\cite{odena2016semi, kumar2017improved, salimans2016improved}, clustering~\cite{springenberg2015unsupervised}, cross domain learning~\cite{taigman2016unsupervised, kim2017learning}, and ensemble learning~\cite{tolstikhin2017adagan}.

\begin{figure}[t!]
\centering
\includegraphics[width=0.75\textwidth]{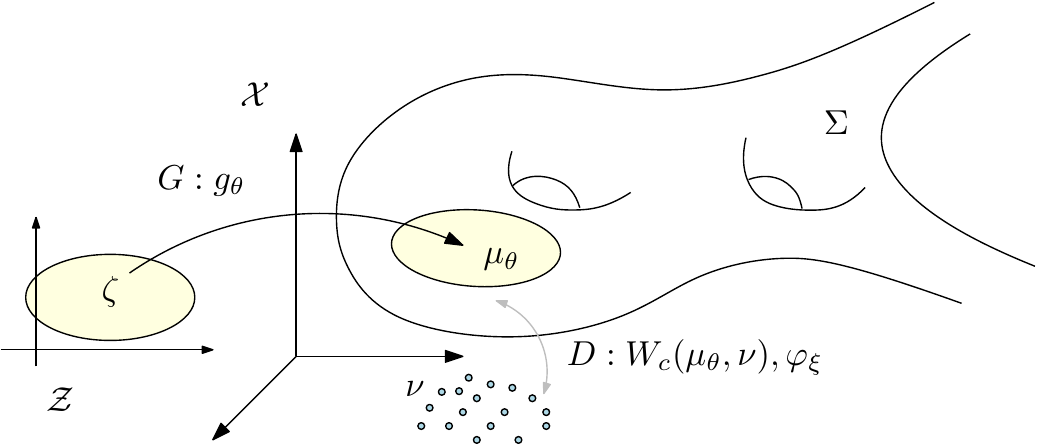}
\caption{Wasserstein Generative Adversarial Networks (W-GAN) framework.
\label{fig:GAN_framework}}
\end{figure}

\paragraph{Optimal Transportation View}
Recently, optimal mass transportation theory has been applied to improve GANs. The Wasserstein distance has been adapted by GANs as the loss function as the discriminator, such as  WGAN~\cite{arjovsky2017wasserstein}, WGAN-GP~\cite{gulrajani2017improved} and RWGAN~\cite{guo2017relaxed}. When the supports of two distributions have no overlap, Wasserstein distance still provides a suitable gradient for the generator to update.

Figure ~\ref{fig:GAN_framework} shows the optimal mass transportation point of view of WGAN ~\cite{arjovsky2017wasserstein}. The ambient image space is $\mathcal{X}$, the real data distribution is $\nu$. The latent space is $\mathcal{Z}$ with much lower dimension. The generator $G$ can be treated as a mapping from the latent space to the sample space, $g_\theta:\mathcal{Z}\to \mathcal{X}$, realized by a deep network with parameter $\theta$. Let $\zeta$ be a fixed distribution on the latent space, such as uniform distribution of Gaussian distribution. The generator $G$ pushes forward $\zeta$ to a distribution $\mu_\theta=g_{\theta\#}\zeta$ in the ambient space $\mathcal{X}$. The discriminator $D$ computes the distance between $\mu_\theta$ and $\nu$, in general using the Wasserstein distance, $W_c(\mu_\theta,\nu)$. The Wasserstein distance is equivalent to find the so-called Kantorovich potential function $\varphi_\xi$, which is carried out by another deep network with parameter $\xi$. Therefore, $G$ improves the ''decoding'' map $g_\theta$ to approximate $\nu$ by $g_{\theta\#}\zeta$; $D$ improves the $\varphi_\xi$ to increase the approximation accuracy to the Wasserstein distance. The generator $G$ and the discriminator $D$ are trained alternatively, until the competition reaches an equilibrium.

In summary, the generative model has natural connection with the optimal mass transportation (OMT) theory:
\begin{enumerate}
\item In generator $G$, the generating map $g_\theta$ in GAN is equivalent to the optimal transportation map in OMT;
\item In discriminator $D$, the metric between distributions is equivalent to the Kantorovich potential $\varphi_\xi$.
\item The alternative training process of W-GAN is the min-max optimization of expectations:
\[
    \min_\theta \max_{\xi} \mathbb{E}_{z\sim \zeta} (\varphi_\xi(g_\theta(z))) + \mathbb{E}_{y\sim \nu}(\varphi_\xi^c(y)).
\]
The deep nets of $D$ and $G$ perform the maximization and the minimization respectively.
\end{enumerate}

\begin{figure}[h!]
\centering
\includegraphics[width=0.75\textwidth]{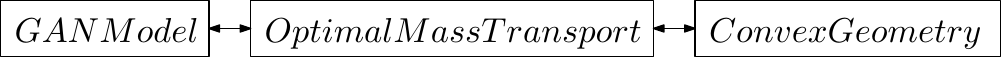}
\caption{The GAN model, OMT theory and convex geometry has intrinsic relations.
\label{fig:GAN_OMT_Alexandrov}}
\end{figure}
\paragraph{Geometric Interpretation} The optimal mass transportation theory has intrinsic connections with the convex geometry. Special OMT problem is equivalent to the Alexandrov theory in convex geometry: finding the optimal transportation map with $L^2$ cost is equivalent to constructing a convex polytope with user prescribed normals and face volumes. The geometric view leads to a practical algorithm, which finds the generating map $g_\theta$ by a convex optimization. Furthermore, the optimization can be carried out using Newton's method with explicit geometric meaning. The geometric interpretation also gives the direct relation between the transportation map $g_\theta$ for $G$ and the Kantorovich potential $\varphi_\xi$ for $D$.

These concepts can be explained using the plain language in computational geometry \cite{Edelsbrunner1987},
\begin{enumerate}
\item the Kantorovich potential $\varphi_\xi$ corresponds to the power distance;
\item the optimal transportation map $g_\theta$ represents the mapping from the power diagram to the power centers, each power cell is mapped to the corresponding site.
\end{enumerate}

\paragraph{Imaginary Adversary} In the current work, we use optimal mass transportation theory to show the fact that: by carefully designing the model and choosing special distance functions $c$, the generator map $g_\theta$ and the descriminator function (Kantorovich potential) $\varphi_\xi$ are equivalent, one can be deduced from the other by a simple closed formula. Therefore, once the Kantorovich potential reaches the optimum, the generator map can be obtained directly without training. One of the deep neural net for $G$ or $D$ is redundant, one of the training processes is wasteful. The competition between the generator $G$ and the discriminator $D$ is unnecessary. In one word, the adversary is imaginary.

\paragraph{Contributions} The major contributions of the current work are as follows:
\begin{enumerate}
\item Give an explicit geometric interpretation of optimal mass transportation map, and apply it for generative model;
\item Prove in theorem \ref{thm:DG_relation} that if the cost function $c(x,y)=h(x-y)$, where $h$ is a strictly convex function, then once the optimal discriminator is obtained, the generator can be written down in an explicit formula. In this section, the competition between the discriminator and the generator is unnecessary and the computational architecture can be simplified;
\item Propose a novel framework for generative model, which uses geometric construction of the optimal mass transportation map;
\item Conduct preliminary experiments for the proof of concepts.
\end{enumerate}

\paragraph{Organization} The article is organized as follows: section \ref{sec:OMT_GAN} explains the optimal transportation view of WGAN in details; section \ref{sec:OMT} lists the main theory of OMT; section \ref{sec:convex_geometry} gives the detailed exposition of Minkowski and Alexandrov theorems in convex geometry, and its close relation with power diagram theory in computational geometry, an explicit computational algorithm is given to solve Alexandrov's problem; section \ref{sec:semi_discrete_OMT} analyzes semi-discrete optimal transportation problem, and connects Alexandrov problem with the optimal transportation map; section \ref{sec:GGM} proposes a novel geometric generative model, which applies the geometric OMT map to the generative model; preliminary experiments are conducted for proof of concept, which are reported in section \ref{sec:experiments}. The work concludes in the section \ref{sec:discussion_conclusion}.

\section{Optimal Transportation View of GAN}
\label{sec:OMT_GAN}
This section, the GAN model is interpreted from the optimal transportation point of view. We show that the discriminator mainly looks for the Kantorovich potential.

Let $\mathcal{X}\subset\mathbb{R}^n$ be the (abient) image space, $\mathcal{P}(\mathcal{X})$ be the Wasserstein space of all probability measures on $\mathcal{X}$. Assume the data distribution is $\nu\in \mathcal{P}(\mathcal{X})$, represented as an empirical distribution
\begin{equation}
    \nu := \frac{1}{n} \sum_{j=1}^n \delta_{y_j},
\end{equation}
where $y_j\in \mathcal{X}, j=1,\dots,n$ are data samples. A \emph{generative model} produces a parametric family of probability distributions $\mu_\theta$, $\theta\in \Theta$, a Minimum Kantorovitch Estimator for $\theta$ is defined as any solution to the problem
\[
    \min_\theta W_c(\mu_\theta,\nu),
\]
where $W_c$ is the Wasserstein cost on $\mathcal{P}(\mathcal{X})$ for some ground cost function $c:\mathcal{X}\times\mathcal{X}\to\mathbb{R}$,
\begin{equation}
    W_c(\mu,\nu)=\min_{\gamma\in \mathcal{P}(\mathcal{X}\times\mathcal{X})}\left\{
    \int_{\mathcal{X}\times\mathcal{X}} c(x,y) d\gamma(x,y)| \pi_{x\#}\gamma=\mu,
    \pi_{y\#}\gamma=\nu
    \right\}
\end{equation}
where $\pi_x$ and $\pi_y$ are projectors, $\pi_{x\#}$ and $\pi_{y\#}$ are marginalization operators. In a generative model, the image samples are encoded to a low dimensional latent space (or a feature space) $\mathcal{Z}\subset \mathbb{R}^m$, $m\ll n$. Let $\zeta$ be a fixed distribution supported on $\mathcal{Z}$. A WGAN produces a parametric mapping $g_\theta: \mathcal{Z}\to \mathcal{X}$, which is treated as a ''decoding'' map the latent space $\mathcal{Z}$ to the original image space $\mathcal{X}$. $g_\theta$ pushes $\zeta$ forward to $\mu_\theta \in \mathcal{P}(\mathcal{X})$, $\mu_\theta = g_{\theta\#}\zeta$. The minimal Kantorovich estimator in WGAN is formulated as
\[
    \min_\theta E(\theta) := W_c( g_{\theta\#}\zeta,\nu).
\]
According to the optimal transportation theory, the Kantorovich problem has a dual formulation
\begin{equation}
    E(\theta) = \max_{\varphi,\psi} \left\{
    \int_{\mathcal{Z}} \varphi(g_\theta(z)) d\zeta(z) + \int_{\mathcal{X}} \psi(y)d\nu(y); \varphi(x)+\psi(y) \le c(x,y)
    \right\}
    \label{eqn:dual_energy}
\end{equation}
The gradient of the dual energy with respect to $\theta$ can be written as
\[
    \nabla E(\theta) = \int_{\mathcal{Z}} [\partial_\theta g_\theta(z)]^T \nabla \varphi^{\star}(g_\theta(z)) d\zeta(z),
\]
where $\varphi^{\star}$ is the optimal Kantorovich potental. In practice, $\psi$ can be replaced by the c-tranform of $\varphi$, defined as
\[
    \varphi^c(y):= \inf_x c(x,y)-\varphi(x).
\]
The function $\varphi$ is called the \emph{Kantorovich potential}. Since $\nu$ is discrete, one can replace the continuous potential
$\varphi^c$ by a discrete vector $\sum_i \psi_i\delta_{y_i}$ and
impose $\varphi = (\sum_i \psi_i\delta_{y_i})^c$. The optimization over $\{\psi_i\}$ can then be achieved using stochastic gradient descent, as in
~\cite{Genevay2016}.

In WGAN~\cite{arjovsky2017wasserstein}, the dual problem Eqn.~\ref{eqn:dual_energy} is solved by approximating the
Kantorovich potential $\varphi$ by the so-called ''adversarial'' map£¬ $\varphi_\xi :\mathcal{X}\to\mathbb{R}$, where $\xi$ is represented by a discriminative deep network. This leads to the Wasserstein-GAN problem
\begin{equation}
    \min_\theta\max_\xi \int_{\mathcal{Z}} \varphi_\xi \circ g_\theta(z) d\zeta(z) +  \frac{1}{n}\sum_{j=1}^n \varphi_\xi^c(y_j).
    \label{eqn:WGAN}
\end{equation}
The generator produces $g_\theta$, the discriminator estimates $\varphi_\xi$, by simultaneous training, the competition reaches the equilibrium.
In WGAN~\cite{arjovsky2017wasserstein}, $c(x,y)=|x-y|$, then the c-transform of $\varphi_\xi$ equals to $-\varphi_\xi$, subject to $\varphi_\xi$ being a 1-Lipschitz function. This is used in  to replace $\varphi_\xi^c$ by $-\varphi_\xi$ in Eqn.~\ref{eqn:WGAN} and use deep network made of ReLu units whose Lipschitz constant is upper-bounded by $1$.
\section{Optimal Mass Transport Theory}
\label{sec:OMT}
In this section, we review the classical optimal mass transportation theory. Theorem \ref{thm:DG_relation} shows the intrinsic relation between the Wasserstein distance (Kantorovich potential ) and the optimal transportation map (Brenier potential), this demonstrates that once the optimal discriminator is known, the optimal generator is automatically obtained. The game between the discriminator and the generator is unnecessary.

The problem of finding a map  that minimizes the inter-domain transportation cost while preserves measure quantities was first studied by Monge \cite{Monge} in the 18th century. Let $X$ and $Y$ be two metric spaces with probability measures $\mu$ and $\nu$ respectively. Assume $X$ and $Y$ have equal total measure
\[
    \int_X d\mu = \int_Y d\nu.
\]
\begin{definition}[Measure-Preserving Map]
A map $T: X\to Y$ is \emph{measure preserving} if for any measurable set $B\subset Y$,
\begin{equation}
    \mu(T^{-1}(B)) = \nu(B).
    \label{eqn:area_preserving_mapping}
\end{equation}
If this condition is satisfied, $\nu$ is said to be the push-forward of $\mu$ by $T$, and we write $\nu = T_\# \mu$.
\end{definition}
If the mapping $T:X\to Y$ is differentiable, then measure-preserving condition can be formulated as the following Jacobian equation, $\mu(x)dx = \nu(T(x)) dT(x)$,
\begin{equation}
   det(DT(x))= \frac{\mu(x)}{\nu\circ T(x)}.
\end{equation}

Let us denote the transportation cost for sending $x\in X$ to $y\in Y$ by $c(x,y)$, then the total \emph{transportation cost} is given by
\begin{equation}
    \mathcal{C}(T):=\int_X c(x,T(x)) d\mu(x).
    \label{eqn:transportation_cost}
\end{equation}

\begin{problem}[Monge's Optimal Mass Transport\cite{Monge}] Given a transportation cost function $c: X\times Y\to \mathbb{R}$, find the measure preserving map $T:X\to Y$ that minimizes the total transportation cost
\begin{equation}
(MP)\hspace{2cm}    W_c(\mu,\nu)= \min_{T:X\to Y} \left\{\int_X c(x,T(x)) d\mu(x):T_\#\mu=\nu \right\}.
    \label{eqn:MP}
\end{equation}
\end{problem}
The total transportation cost $W_c(\mu,\nu)$ is called the \emph{Wasserstein distance} between the two measures $\mu$ and $\nu$.

\subsection{Kantorovich's Approach }

In the 1940s, Kantorovich introduced the relaxation of Monge's problem~ \cite{Kantorovich48}. Any strategy for sending $\mu$ onto $\nu$ can be represented by a joint measure $\rho$ on $X\times Y$, such that
\begin{equation}
    \rho(A\times Y) = \mu(A), \rho(X\times B) = \nu(B),
    \label{eqn:marginal_measure}
\end{equation}
$\rho(A\times B)$ is called a \emph{transportation
plan}, which represents the share to be moved from $A$ to $B$. We denote the projection to $X$ and $Y$ as  $\pi_x$ and $\pi_y$ respectively, then $\pi_{x\#} \rho = \mu$ and $\pi_{y\#} \rho = \nu$. The total cost of the transportation plan $\rho$ is
\begin{equation}
    \mathcal{C}(\rho):= \int_{X\times Y} c(x,y) d\rho(x,y).
    \label{eqn:transportation_cost_2}
\end{equation}
The Monge-Kantorovich problem consists in finding the $\rho$, among all the suitable transportation plans, minimizing $\mathcal{C}(\rho)$ in Eqn.~\ref{eqn:transportation_cost_2}£¬
\begin{equation}
(KP)\hspace{2cm}    W_c(\mu,\nu) := \min_{\rho} \left\{ \int_{X\times Y} c(x,y) d\rho(x,y): \pi_{x\#}\rho = \mu, \pi_{y\#}\rho = \nu \right\}
    \label{eqn:KP}
\end{equation}

\subsection{Kontarovich Dual Formulation}

Because Eqn.~\ref{eqn:KP} is a linear program, it has a dual formulation, known as the Kantorovich problem \cite{villani2008optimal}:
\begin{equation}
(DP)\hspace{2cm}   W_c(\mu,\nu) := \max_{\varphi,\psi} \left\{ \int_{X} \varphi(x) d\mu(x) + \int_Y \psi(y) d\nu(y): \varphi(x)+\psi(y)\le c(x,y) \right\}
    \label{eqn:dual_formulation}
\end{equation}
where $\varphi:X\to\mathbb{R}$ and $\psi:Y\to\mathbb{R}$ are real functions defined on $X$ and $Y$. Equivalently, we can replace $\psi$ by the c-transform of $\varphi$.

\begin{definition}[c-transform] Given a real function $\varphi: X\to \mathbb{R}$, the c-transform of $\varphi$ is defined by
\[
    \varphi^c(y) = \inf_{x\in X} \left(c(x,y) - \varphi(x)\right).
\]
\end{definition}
Then the Kantorovich problem can be reformulated as the following dual problem:
\begin{equation}
(DP)\hspace{2cm}    W_c(\mu,\nu) := \max_{\varphi} \left\{ \int_{X} \varphi(x) d\mu(x) + \int_Y \varphi^c(y) d\nu(y) \right\},
    \label{eqn:DP}
\end{equation}
where $\varphi:X\to\mathbb{R}$ is called the \emph{Kantorovich potential}.

For $L^1$ transportation cost $c(x,y)=|x-y|$ in $\mathbb{R}^n$, if the Kantorovich potential $\varphi$ is 1-Lipsitz, then its c-transform has a special relation $\varphi^c = -\varphi$. The Wasserstein distance is given by
\begin{equation}
    W_c(\mu,\nu) := \max_{\varphi} \left\{ \int_{X} \varphi(x) d\mu(x) - \int_Y \varphi(y) d\nu(y) \right\},
    \label{eqn:L1DP}
\end{equation}

For $L^2$ transportation cost $c(x,y)=1/2|x-y|^2$ in $\mathbb{R}^n$, the c-transform and the classical Legendre transform has special relations.
\begin{definition} Given a function $\varphi:\mathbb{R}^n\to \mathbb{R}$, its Legendre tranform is defined as
\begin{equation}
    \varphi^*(y):=\sup_x \left(\langle x, y \rangle - \varphi(x)\right).
    \label{eqn:Legendre_transform}
\end{equation}
\end{definition}
Intuitively, Legendre tranform has the following form:
\[
    \left(\int xdy\right)^* = \int ydx.
\]
We can show the following relation holds when $c=1/2|x-y|^2$,
\begin{equation}
    \frac{1}{2} |y|^2 - \varphi^c = \left( \frac{1}{2} |x|^2 - \varphi \right)^*.
\end{equation}

\subsection{Brenier's Approach }

At the end of 1980's, Brenier \cite{brenier} discovered the
intrinsic connection between optimal mass transport map and convex
geometry. (see also for instance \cite{vil}, Theorem 2.12(ii), and Theorem 2.32)

Suppose $u:X\to \mathbb{R}$ is a $C^2$ continuous convex function, namely its Hessian matrix is semi-positive definite. $\left( \partial^2 f/\partial x_i
\partial x_j \right ) \ge  0.$ Its gradient map $\nabla u: X\to Y $is defined as $x\mapsto \nabla u(x).$

\begin{theorem}[Brenier\cite{brenier}] Suppose $X$ and $Y$ are the Euclidean space $\mathbb{R}^n$, and the transportation cost is the quadratic Euclidean distance $c(x,y) = |x-y|^2$. If $\mu$ is absolutely continuous and $\mu$ and $\nu$ have finite second order moments, then there exists a convex function $u: X\to \mathbb{R}$, its gradient map $ \nabla u$ gives the solution to the Monge's problem, where $u$ is called Brenier's potential. Furthermore, the optimal mass transportation map is unique.
\label{thm:Brenier}
\end{theorem}
This theorem converts the Monge's problem to solving the following Monge-Amper\'e partial differential equation:
\begin{equation}
det \left(
\frac{\partial^2 u}{\partial x_i \partial x_j}\right )(x) = \frac{\mu(x)}{\nu
\circ \nabla u(x)}.
\end{equation}
The function $u:X\to\mathbb{R}$ is called the \emph{Brenier potential}. Brenier proved the polar factorization theorem.

\begin{theorem}[Brenier Factorization\cite{brenier}] Suppose $X$ and $Y$ are the Euclidean space $\mathbb{R}^n$, $\varphi:X\to Y$ is measure preserving, $\varphi_\# \mu = \nu$. Then there exists a convex function $u:X\to\mathbb{R}$, such that
\[
    \varphi = \nabla u\circ s,
\]
where $s:X\to X$ preserves the measure $\mu$, $s_\# \mu = \mu$. Furthermore, this factorization is unique.
\label{thm:Brenier_factorization}
\end{theorem}

Based on the generalized Brenier theorem we can obtain the following theorem.

\begin{theorem}[Generator-Discriminator Equivalence]Given $\mu$ and $\nu$ on a compact domain $\Omega\subset \mathbb{R}^n$ there exists an optimal transport plan $\gamma$ for the cost $c(x,y)=h(x-y)$ with $h$ strictly convex. It is unique and of the form $(id,T_\#)\mu$, provided $\mu$ is absolutely continuous and $\partial \Omega$ is negligible. More over, there exists a Kantorovich potential $\varphi$, and $T$ can be represented as
\[
    T(x)= x- (\nabla h)^{-1}(\nabla \varphi(x)).
\]
\label{thm:DG_relation}
\end{theorem}
\noindent{Proof:} Assume $\rho$ is the joint probability, satisfying the conditions $\pi_{x\#}\rho = \mu$, $\pi_{y\#}\rho=\nu$, $(x_0,y_0)$ is a point in the support of $\rho$, by definition $\varphi^c(y_0) = \inf_x c(x,y_0) - \varphi(x)$, hence
\[
    \nabla \varphi(x_0) = \nabla_x c(x_0,y_0)=\nabla h(x_0-y_0),
\]
Because $h$ is strictly convex, therefore $\nabla h$ is invertible,
\[
    x_0 - y_0 = (\nabla h)^{-1}( \nabla \varphi(x_0)),
\]
hence $y_0 = x_0 - (\nabla h)^{-1}( \nabla \varphi(x_0))$. $\square$\\
\\
When $c(x,y)=\frac{1}{2}|x-y|^2$, we have
\[
    T(x)=x-\nabla \varphi(x) = \nabla \left( \frac{x^2}{2}-\varphi(x) \right) = \nabla u(x).
\]
In this case, the Brenier's potential $u$ and the Kantorovich's potential $\varphi$ is related by
\begin{equation}
    u(x) = \frac{x^2}{2}-\varphi(x).
\end{equation}
\section{Convex Geometry}
\label{sec:convex_geometry}
\begin{figure}[h]
\centering
\begin{tabular}{cc}
\includegraphics[width=0.3\textwidth]{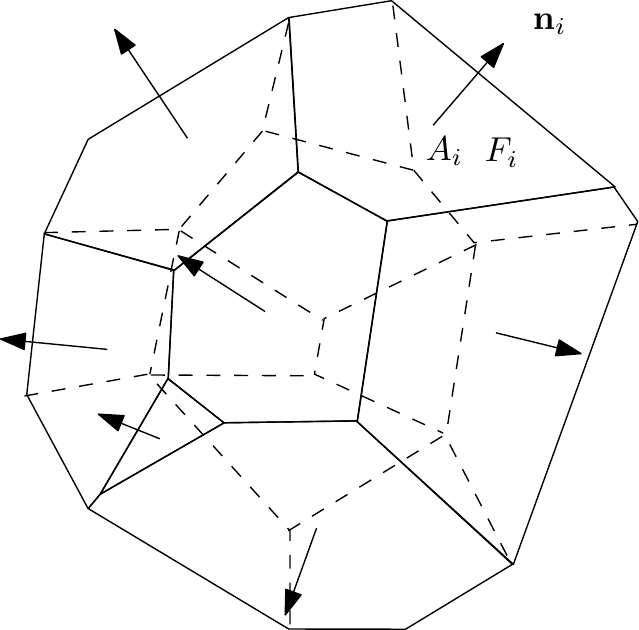}&
\includegraphics[width=0.4\textwidth]{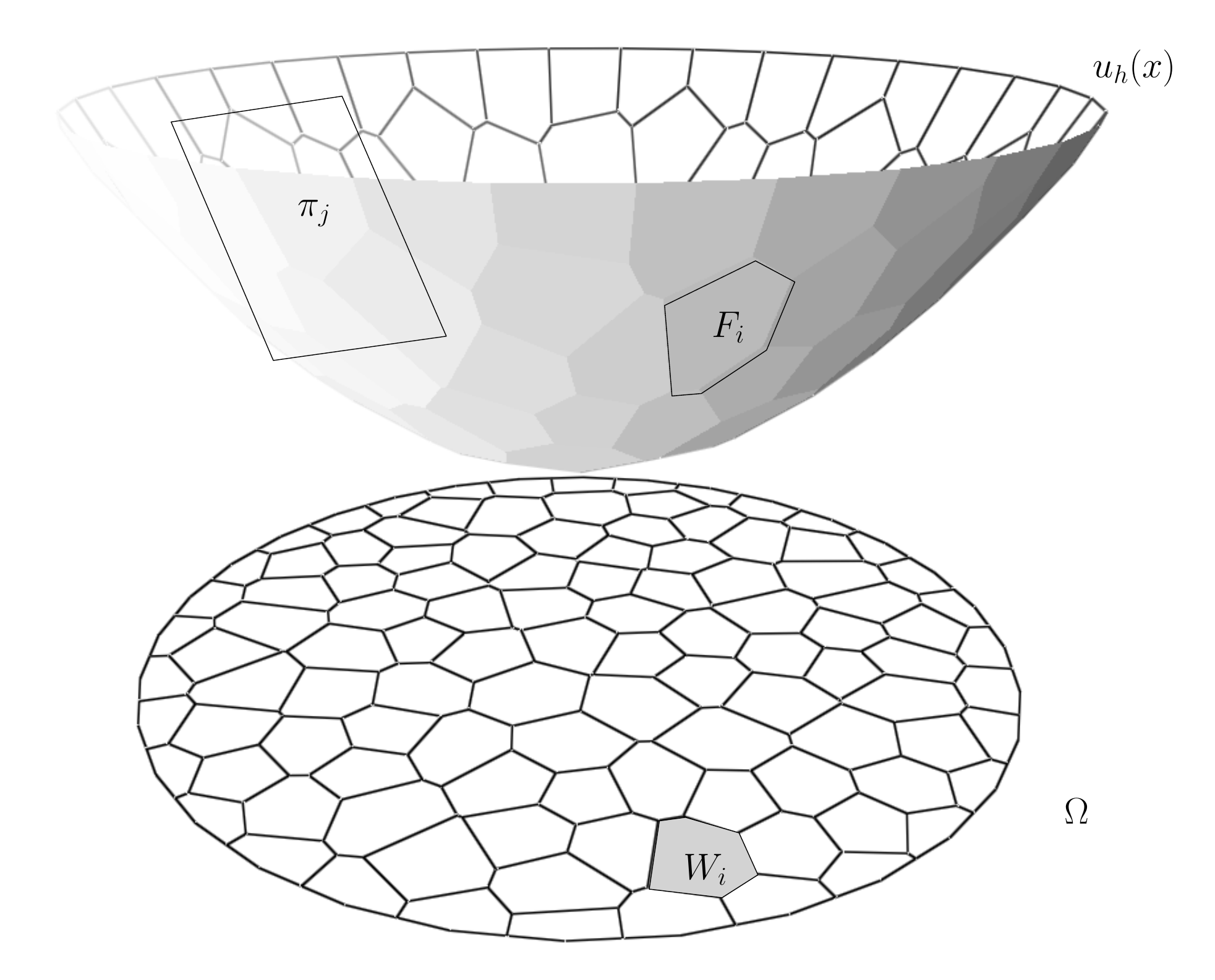}\\
(a) Minkowski theorem &(b) Alexandrov theorem
\end{tabular}
\caption{Minkowski and Alexandrov theorems for convex polytopes with prescribed normals and areas.
\label{fig:minkowski_problem}}
\end{figure}
This section introduces Minkowski and Alexandrov problems in convex geometry, which can be described by Monge-Ampere equation as well. This intrinsic connection gives a geometric interpretation to optimal mass transportation map with $L^2$ transportation cost.

\subsection{Alexandrov' Theorem}
Minkowski proved the existence and the uniqueness of convex polytope with user prescribed face normals and the areas.
\begin{theorem}[Minkowski] Suppose $n_1, ..., n_k$ are unit vectors which
span $\mathbb{R}^n$ and $\nu_1, ..., \nu_k>0$ so that $\sum_{i=1}^k \nu_i
n_i=0$. There exists a compact convex polytope $P \subset \mathbb{R}^n$ with
exactly $k$ codimension-1 faces $F_1, ..., F_k$ so that $n_i$ is
the outward normal vector to $F_i$ and the volume of $F_i$ is $\nu_i$. Furthermore, such $P$ is unique up to parallel translation.
\end{theorem}

Minkowski's proof is variational and suggests an algorithm to find
the polytope. Minkowski theorem for unbounded convex polytopes was considered
and solved by A.D. Alexandrov and his student A. Pogorelov. In his book on convex polyhedra \cite{alex}, Alexandrov proved the following fundamental theorem (Theorem 7.3.2 and theorem 6.4.2)

\begin{theorem} [Alexandrov\cite{alex}]
Suppose $\Omega$ is a compact convex polytope with non-empty
interior in $\mathbb{R}^n$, $n_1, ..., n_k \subset \mathbb{R}^{n+1}$ are
distinct $k$ unit vectors, the $(n+1)$-th coordinates are negative, and $\nu_1, ..., \nu_k >0$ so that $\sum_{i=1}^k \nu_i=vol(\Omega)$. Then there exists convex polytope $P\subset \mathbb{R}^{n+1}$ with exact $k$ codimension-1 faces$F_1,\dots,F_k$£¬ so that $n_i$ is the normal vector to $F_i$ and the intersection between $\Omega$ and the projection of $F_i$ is with volume $\nu_i$. Furthermore, such $P$ is unique up to vertical translation.
\end{theorem}


Alexandrov's proof is based on algebraic topology and non-constructive. Gu et al. \cite{Gu_AJM_2016} gave a variational proof for the generalized Alexandrov theorem stated in terms of
convex functions.

Given $y_1,\dots,y_k\in \mathbb{R}^n$ and $h=(h_1,\dots,h_k)\in\mathbb{R}^k$, the piecewise linear convex function is defined as
\[
    u_h(x) = \max_{i} \left\{\langle x,y_i\rangle + h_i \right\}.
\]
The graph if $u_h$ is a convex polytope in $\mathbb{R}^{n+1}$, the projection induces a cell decomposition of $\mathbb{R}^n$¡£ Each cell is a closed convex polytope,
\[
    W_i(h) = \left\{x\in \mathbb{R}^n | \nabla u_h(x) = y_i \right\}.
\]
Some cells may be empty or unbounded. Given a probability measure $\mu$ defined on $\Omega$, the volume of $W_i(h)$ is defined as
\[
     w_i(h):= \mu(W_i(h)\cap\Omega)=\int_{W_i(h) \cap \Omega} d\mu.
\]

\begin{theorem}[Gu-Luo-Sun-Yau\cite{Gu_AJM_2016}]  Let $\Omega$ be a compact convex domain in $\mathbb{R}^n$,
 $\{y_1, ..., y_k\}$ be a set of distinct points in $\mathbb{R}^n$ and
$\mu$ a probability measure on $\Omega$. Then for any $\nu_1, ..., \nu_k >0$ with $\sum_{i=1}^k \nu_i =\mu(\Omega)$, there exists $h=(h_1, ..., h_k) \in \mathbb{R}^k$,
unique up to adding a constant $(c,..., c)$, so that
$w_i(h)=\nu_i$, for all $i$. The vectors $h$ are exactly maximum points of the concave function
\begin{equation}
    E(h) = \sum_{i=1}^k h_i \nu_i-\int^h_0 \sum_{i=1}^k w_i(\eta) d\eta_i
\label{eqn:Alexandrov_energy}
\end{equation}
on the open convex set
\[
    H =\{ h \in \mathbb{R}^k | w_i(h) >0, \forall i\}.
\]
Furthermore,
$\nabla u_h$ minimizes the quadratic cost
\[
    \int_{\Omega} | x - T(x)|^2 d\mu(x)
\]
among all transport maps $T_\# \mu=\nu$, where the Dirac measure $\nu=\sum_{i=1}^k \nu_i \delta_{y_i}$.
\label{thm:otp}
\end{theorem}

For the convenience of discussion, we define the Alexandrov's potential as follows:
\begin{definition}[Alexandrov Potential] Under the above condition, the convex function
\begin{equation}
    \mathcal{A}(h) = \int^h \sum_{i=1}^k w_i(\eta) d\eta_i
\label{eqn:Alexandrov_potential}
\end{equation}
is called the Alexandrov potential.
\end{definition}

\begin{figure}[t!]
\centering
\includegraphics[width=0.8\textwidth]{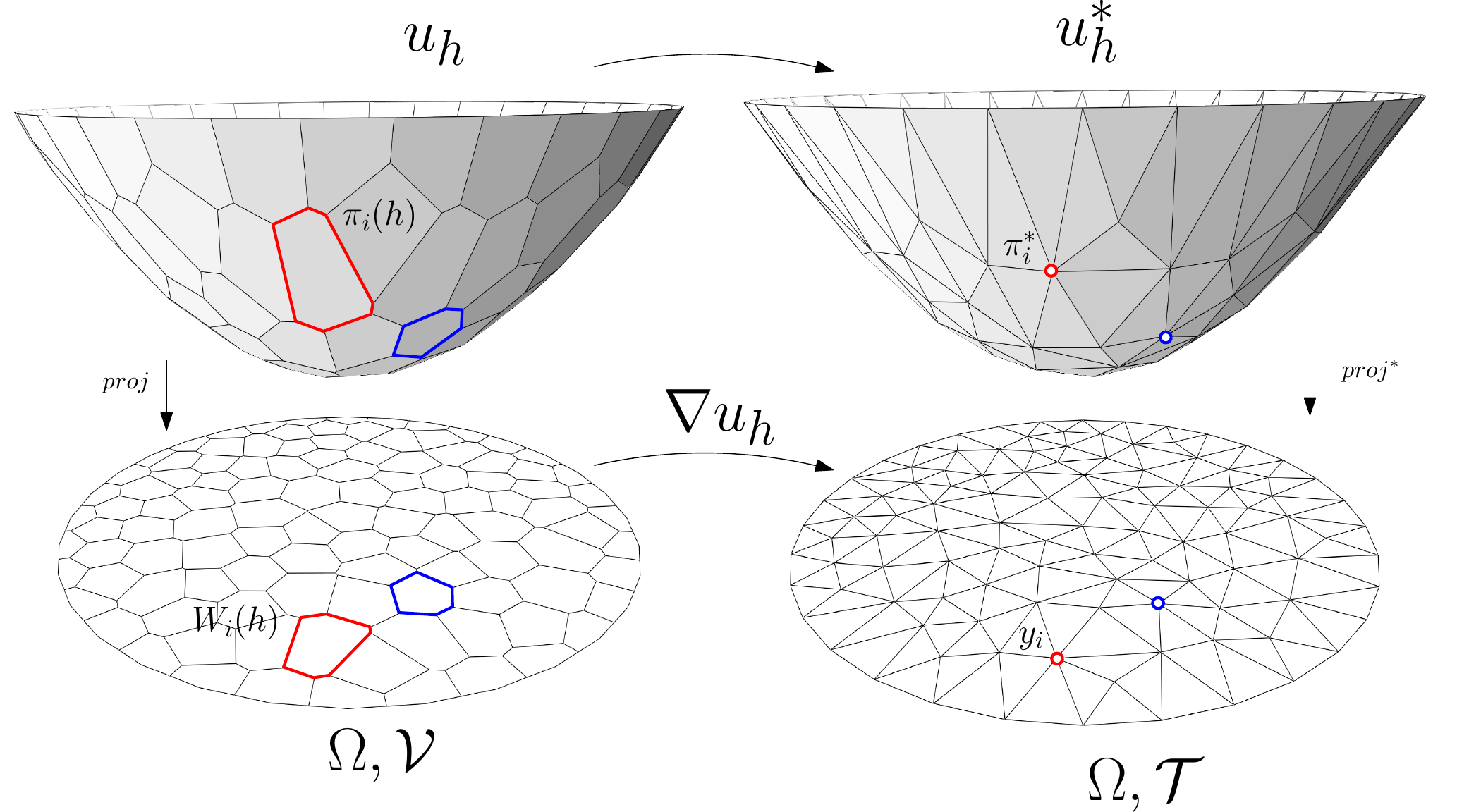}
\caption{Geometric Interpretation to Optimal Transport Map: Brenier potential $u_h:\Omega\to\mathbb{R}$, Legendre dual $u_h^*$, optimal transportation map $\nabla u_h:W_i(h)\to y_i$, power diagram $\mathcal{V}$, weighted Delaunay triangulation $\mathcal{T}$.
\label{fig:discrete_VP_framework}}
\end{figure}

\begin{figure}[!t]
\centering
\includegraphics[width=0.75\textwidth]{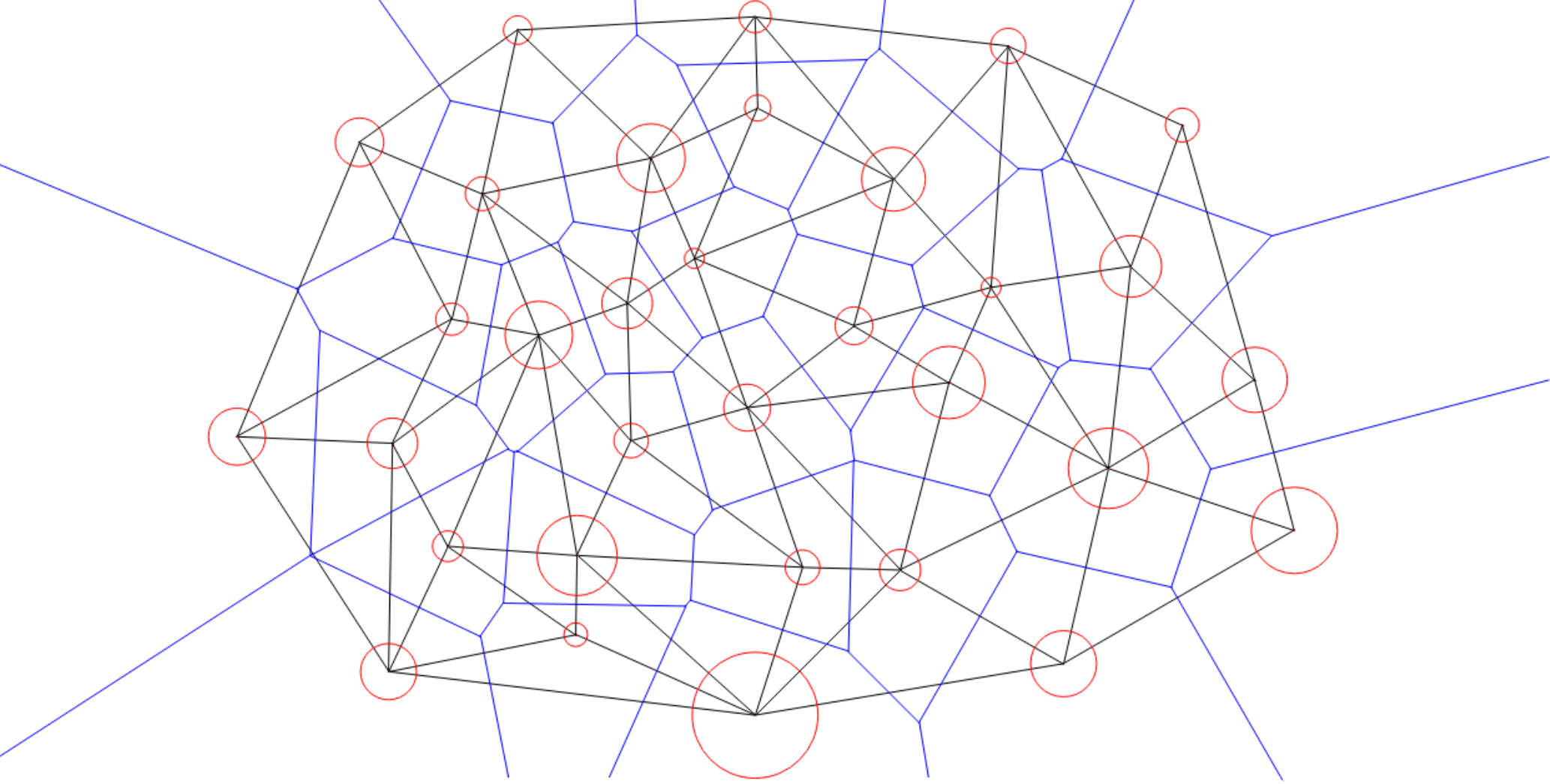}
\caption{Power diagram (blue) and its dual weighted Delaunay triangulation (black), the power weight $\psi_i$ equal to the square of radius $r_i$ (red circle).
\label{fig:power_diagram}}
\end{figure}

\subsection{Power Diagram}

Alexandrov's theorem has close relation with the conventional power diagram. We can use power diagram algorithm to solve the Alexandrov's problem.

\begin{definition}[power distance]Given a point $y_i\in \mathbb{R}^n$ with a power weight $\psi_i$, the power distance is given by
\[
    \text{pow}(x,y_i) = |x-y_i|^2 - \psi_i.
\]
\end{definition}
\begin{definition}[power diagram]
Given weighted points $\{(y_1,\psi_1),(y_2,\psi_2),\dots, (y_k,\psi_k)\}$,  the power diagram is the cell decomposition of $\mathbb{R}^n$, denoted as $\mathcal{V}(\psi)$,
\[
    \mathbb{R}^n = \bigcup_{i=1}^k W_i(\psi),
\]
where each cell is a convex polytope
\[
    W_i(\psi) = \{x\in \mathbb{R}^n | \text{pow}(x,y_i) \le \text{pow}(x,y_j),\forall j\}.
\]
\end{definition}
The weighted Delaunay triangulation, denoted as $\mathcal{T}(\psi)$, is the Poincar\'e dual to the power diagram, if $W_i(\psi)\cap W_j(\psi)\neq \emptyset$ then there is an edge connecting $y_i$ and $y_j$ in the weighted Delaunay triangulation.\\

Note that $\text{pow}(x,y_i) \le \text{pow}(x,y_j)$ is equivalent to
\[
    \langle x,y_i \rangle + \frac{1}{2}(\psi_i - |y_i|^2) \ge \langle x,y_j \rangle + \frac{1}{2}(\psi_j - |y_j|^2).
\]
let
\begin{equation}
h_i = 1/2(\psi_i - |y_i|^2),
\end{equation}
we construct the convex function
\begin{equation}
    u_h(x) = \max_{i} \{ \langle x,y_i \rangle + h_i \}.
\end{equation}

\subsection{Convex Optimization}

Now, we can use the power diagram to explain the gradient and the Hessian of the energy Eqn.\ref{eqn:Alexandrov_energy}, by definition
\begin{equation}
    \nabla E(h) = (\nu_1 - w_1(h), \nu_2 - w_2(h), \cdots, \nu_k - w_k(h))^T.
\label{eqn:Alexandrov_gradient}
\end{equation}
The Hessian matrix is given by power diagram - weighted Delaunay triangulation, for adjacent cells in the power diagram,
\begin{equation}
\frac{\partial^2 E(h)}{\partial h_i\partial h_j} = \frac{\partial w_i(h)}{\partial h_j} = - \frac{\mu(W_i(h)\cap W_j(h) \cap \Omega)}{|y_j-y_i|}
\end{equation}
Suppose edge $e_{ij}$ is in the weighted Delaunay triangulation, connecting $y_i$ and $y_j$. It has a unique dual cell $D_{ij}$ in the power diagram, then
\[
    \frac{\partial w_i(h)}{\partial h_j} = - \frac{\mu(D_{ij})}{|e_{ij}|},
\]
the volume ratio between the dual cells. The diagonal element in the Hessian is
\begin{equation}
\frac{\partial^2 E(h)}{\partial h_i^2} = \frac{\partial w_i(h)}{\partial h_i} = \sum_{j\neq i} \frac{\partial w_i(h)}{\partial h_j}.
\end{equation}
Therefore, in order to solve Alexandrov's problem to construct the convex polytope with user prescribed normal and face volume, we can optimize the energy in Eqn.~\ref{eqn:Alexandrov_energy} using classical Newton's method directly.

Let's observe the convex function $u_h^*$, its graph is the convex hull $\mathcal{C}(h)$.
Then the discrete Hessian determinant of $u_h^*$ assigns each vertex $v$ of $\mathcal{C}(h)$ the volume of the convex hull of the gradients of $u_h^*$ at top-dimensional cells adjacent to $v$. Therefore, solving Alexandrov's problem is equivalent to solve a discrete Monge-Ampere equation.
\section{Semi-discrete Optimal Mass Transport}
\label{sec:semi_discrete_OMT}
In this section, we solve the semi-discrete optimal transportation problem from geometric point of view. This special case is useful in practice.

Suppose $\mu$ has compact support $\Omega$ on $X$, assume $\Omega$ is a convex domain in $X$,
\[
    \Omega = \text{supp}~\mu = \{x\in X| \mu(x) > 0\}.
\]
The space $Y$ is discretized to $Y=\{y_1,y_2,\cdots,y_k\}$ with Dirac measure $\nu =
\sum_{j=1}^k \nu_j \delta(y-y_j)$. The total mass are equal
\[
    \int_{\Omega} d\mu(x) = \sum_{i=1}^k \nu_i.
\]

\subsection{Kantorovich Dual Approach}

%

We define the discrete Kantorovich potential $\psi: Y\to \mathbb{R}$, $\psi(y_j) = \psi_j$, then
\begin{equation}
    \int_Y \psi d\nu = \sum_{j=1}^k \psi_j \nu_j.
    \label{eqn:nu_expectation}
\end{equation}
The c-transformation of $\psi$ is given by
\begin{equation}
    \psi^c(x) = \min_{1\le j \le k} \{ c(x,y_j)-\psi_j\}.
\end{equation}
This induces a cell decomposition of $X$,
\[
    X = \bigcup_{i=1}^k W_i(\psi),
\]
where each cell is given by
\[
    W_i(\psi) = \left\{x\in X| c(x,y_i) - \psi_i \le c(x,y_j) - \psi_j, \forall 1\le j\le k\right\}.
\]


According to the dual formulation of the Wasserstein distance Eqn.\ref{eqn:DP} and integration Eqn.\ref{eqn:nu_expectation}, we define the energy
\[
     E(\psi) = \int_X \psi^c d\mu + \int_Y \psi d\nu
\]
then obtain the formula
\begin{equation}
     E_D(\psi) = \sum_{i=1}^k \psi_i \left(\nu_i - w_i(\psi)\right) + \sum_{j=1}^k \int_{W_j(\psi)} c(x,y_j) d\mu.
\end{equation}
where $w_i(\psi)$ is the measure of the cell $W_i(\psi)$,
\begin{equation}
    w_i(\psi) = \mu(W_i(\psi))=\int_{W_i(\psi)} d\mu(x).
    \label{eqn:power_cell_measure}
\end{equation}
Then the Wasserstein distance between $\mu$ and $\nu$ equals to
\[
    W_c(\mu,\nu) = \max_\psi E(\psi).
\]

\subsection{Brenier's Approach}
Kantorovich's dual approach is for general cost functions. When the cost function is the $L^2$ distance $c(x,y)=|x-y|^2$, we can apply Brenier's approach directly.

We define a \emph{height vector} $h=(h_1,h_2,\cdots,h_k)\in\mathbb{R}^n$, consisting of $k$ real numbers. For each $y_i \in Y$, we construct a hyperplane
defined on $X$, $\pi_i(h):\langle x, y_i\rangle + h_i = 0$. We define the Brenier potential function as
\begin{equation}
    u_h(x) = \max_{i=1}^k \{\langle x, y_i \rangle + h_i\},
    \label{eqn:convex_function}
\end{equation}
then $u_h(x)$ is a convex function. The graph of $u_h(x)$ is an infinite convex polyhedron with supporting planes $\pi_i(h)$. The projection of
the graph induces a polygonal partition of $\Omega$,
\begin{equation}
    \Omega = \bigcup_{i=1}^k W_i(h),
    \label{eqn:cell_decomposition}
\end{equation}
where each cell $W_i(h)$ is the projection of
a facet of the graph of $u_h$ onto $\Omega$,
\begin{equation}
    W_i(h) = \{x\in X | \nabla u_h(x) = y_i \}\cap \Omega.
    \label{eqn:cell}
\end{equation}
The measure of $W_i(h)$ is given by
\begin{equation}
    w_i(h) = \int_{W_i(h)} d\mu.
\label{eqn:cell_area}
\end{equation}
The convex function $u_h$ on each cell
$W_i(h)$ is a linear function $\pi_i(h)$,
therefore, the gradient map
\begin{equation}
\nabla u_h: W_i(h)\to y_i, i = 1, 2, \cdots, k.
\label{eqn:optimal_gradient_map}
\end{equation}
maps each $W_i(h)$ to a single point $y_i$. According to Alexandrov's theorem, and the Gu-Luo-Yau theorem, we obtain the following corollary:
\begin{corollary}Let $\Omega$ be a compact convex domain in $\mathbb{R}^n$,
 $\{y_1, ..., y_k\}$ be a set of distinct points in $\mathbb{R}^n$ and
$\mu$ a probability measure on $\Omega$. Then for any $\nu=\sum_{i=1}^k \nu_i\delta_{y_i}$, with $\sum_{i=1}^k \nu_i =\mu(\Omega)$, there exists $h=(h_1, ..., h_k) \in \mathbb{R}^k$,
unique up to adding a constant $(c,..., c)$, so that
$w_i(h)=\nu_i$, for all $i$. The vectors $h$ are exactly maximum points of the concave function
\begin{equation}
    E_B(h) = \sum_{i=1}^k h_i \nu_i-\int^h_0 \sum_{i=1}^k w_i(\eta) d\eta_i
\label{eqn:Alexandrov_energy_2}
\end{equation}
Furthermore,
$\nabla u_h$ minimizes the quadratic cost
\[
    \int_{\Omega} | x - T(x)|^2 d\mu
\]
among all transport maps $T_\# \mu=\nu$.
\label{thm:otp}
\end{corollary}

\subsection{Equivalence}
For $c(x,y)=1/2|x-y|^2$ cost cases, we have introduced two approaches: Kantorovich's dual approach and Brenier's approach. In the following, we show these two approaches are equivalent.

In Kantorovich's dual approach, finding the optimal mass transportation is equivalent to maximize the following energy:
\[
    E_D(\psi) = \sum_{i=1}^k \psi_i (\nu_i - w_i(\psi)) + \sum_{j=1}^k \int_{W_j(\psi)} c(x,y_j) d\mu.
\]
In Brenier's approach, finding the optimal transportation map boils down to maximize
\[
    E_B(h) = \sum_{i=1}^k h_i \nu_i - \int^h \sum_{i=1}^k w_i(\eta) d\eta.
\]

\begin{lemma}
Let $\Omega$ be a compact convex domain in $\mathbb{R}^n$,
 $\{y_1, ..., y_k\}$ be a set of distinct points in $\mathbb{R}^n$. Given
$\mu$ a probability measure on $\Omega$,  $\nu=\sum_{i=1}^k \nu_i\delta_{y_i}$, with $\sum_{i=1}^k \nu_i =\mu(\Omega)$. If $c(x,y)=1/2|x-y|^2$, then
\[
    h_i = \psi_i - \frac{1}{2} |y_i|^2,~~\forall i
\]
and
\[
    E_D(\psi) - E_B( h) = \text{Const}
\]
\end{lemma}

\begin{figure}[!t]
\centering
\includegraphics[width=0.7\textwidth]{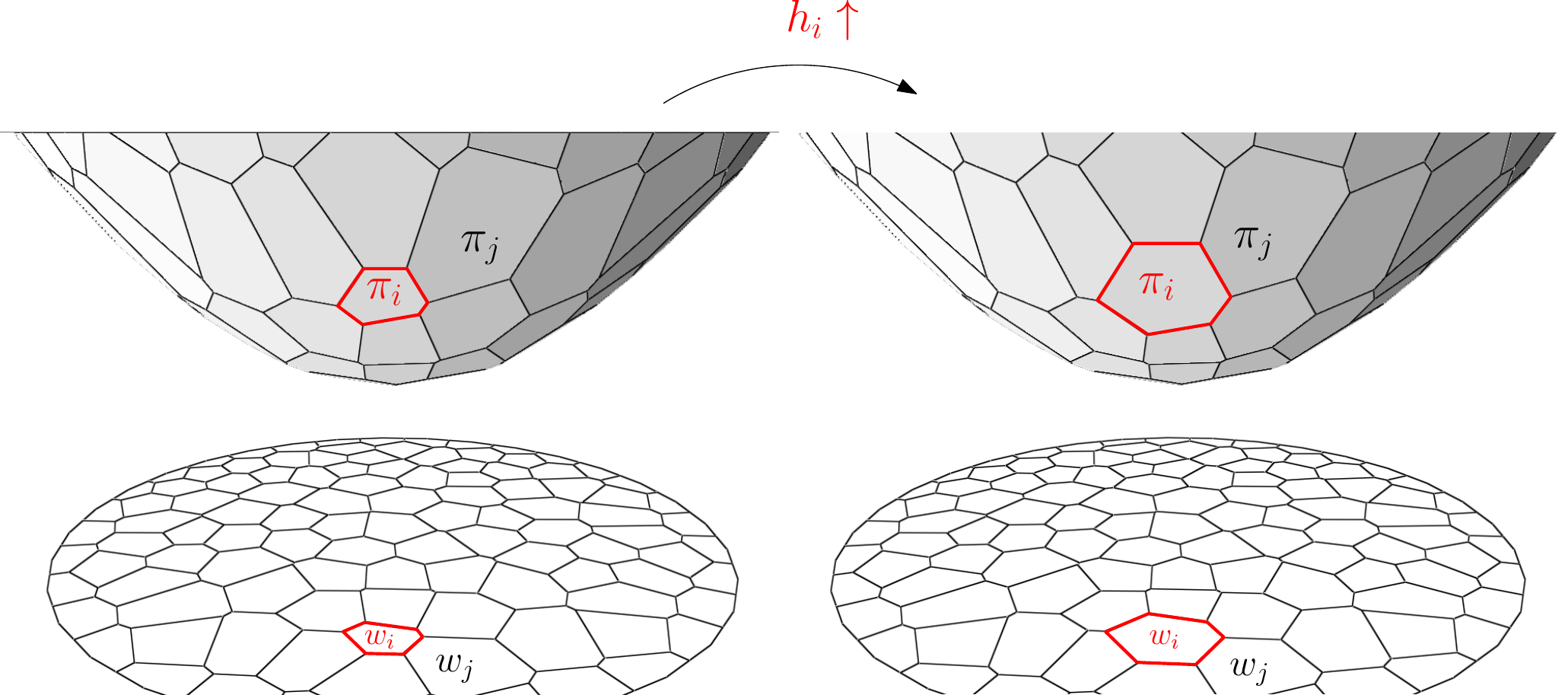}
\caption{Variation of the volume of top-dimensional cells¡£
\label{fig:variation_1}}
\end{figure}

\noindent{proof}: Consider the power cell
\[
    c(x,y_i) - \psi_i \le c(x,y_j) - \psi_j
\]
is equivalent to
\[
    \langle x,y_i \rangle + \left(\psi_i - \frac{1}{2} |y_i|^2 \right)
    \ge \langle x,y_j \rangle + \left(\psi_j - \frac{1}{2} |y_j|^2 \right)
\]
therefore $h_i = \psi_i - 1/2 |y_i|^2$.

Let the transportation cost to be defined as
\[
    \mathcal{C}(\psi) = \sum_{j=1}^k \int_{W_j(\psi)} c(x,y_j) d\mu.
\]
Suppose we infinitesimally change $h$ to $h+dh$, then we define
\[
    D_{ij} = W_j(h) \cap W_i(h+dh) \cap \Omega.
\]
Then $\mu(D_{ij}) = dw_i$, also $\mu(D_{ij}) = -dw_j$. For each $x\in D_{ij}$, $c(x,y_i) - \psi_i =  c(x,y_j) - \psi_j$, then $ c(x,y_i) - c(x,y_j) = \psi_i - \psi_j$, hence
\[
    \int_{D_{ij}} (c(x,y_i) - c(x,y_j) ) d\mu = \psi_i dw_i + \psi_j dw_j.
\]
This shows $d\mathcal{C} = \sum_{i=1}^k \psi_i dw_i$,
hence
\[
    \mathcal{C}(w) = \int^w \sum_{i=1}^k \psi_i dw_i.
\]
The Legendre dual of $\mathcal{C}$ is
\[
   \mathcal{C}^*(\psi)=\int^{\psi} \sum_{i=1}^k w_i d\psi_i.
\]
Hence
\[
\int^w \sum_{i=1}^k \psi_i dw_i + \int^{\psi} \sum_{i=1}^k w_i d\psi_i = \sum_{i=1}^k w_i \psi_i.
\]
On the other hand, $\psi_i = h_i + 1/2|y_i|^2$, $d\psi_i = d h_i$,
\[
    \int^{h} \sum_{i=1}^k w_i dh_i = \int^{\psi} \sum_{i=1}^k w_i d\psi_i + const.
\]
We put everything together
\[
E_D(\psi) - E_B(h) = \sum_i (\psi_i - h_i) \nu_i - \left (\sum_i \psi_i\nu_i - \mathcal{C}(\psi) - \mathcal{C}^*(w)\right ) - c_1 = c_2,
\]
where $C_1$ and $C_2$ are two constants.
$\square$

This shows Kantorovich's dual approach and Brienier's approach are equivalent. At the optimal point, $\nu_i = w_i(\psi)$, therefore $E_D(\psi)$ equals to the transportation cost $\mathcal{C}(\psi)$. Furthermore, the Brenier's potential is
\[
    u_h(x) = \max_{i=1}^k \{\langle x, p_i \rangle + h_i \},
\]
where $h_i$ is given by the power weight $\psi_i$. The Kantorovich's potential is the power distance
\[
    \varphi(x) = \psi^c(x) = \min_j \{c(x,y_j)-\psi_j\} =
    \min_j \{\text{pow}(x,y_j)\} =
    \frac{1}{2}|x|^2 - \max_j \{\langle x, y_j \rangle + (\psi_j - \frac{1}{2} |y_j|^2 )\}
\]
hence at the optimum, the Brenier potential and the Kantorovich potential are related by
\begin{equation}
    u_h(x) = \frac{1}{2}|x|^2-\varphi(x).
    \label{eqn:kantorovich_brenier_potentials}
\end{equation}

\begin{figure}[t!]
\centering
\includegraphics[width=0.75\textwidth]{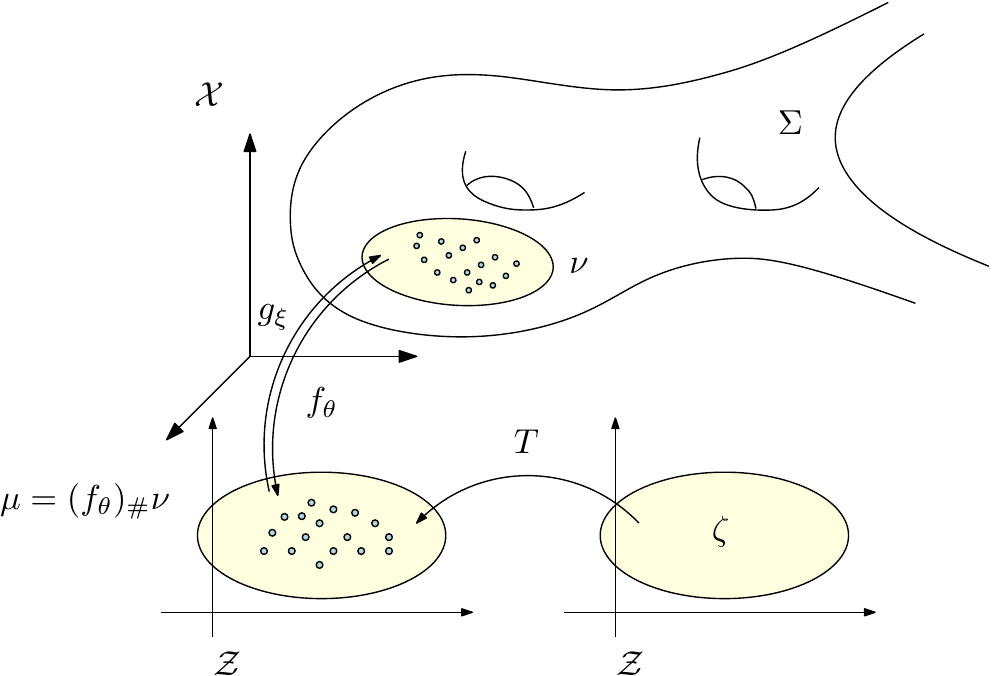}
\caption{The framework for Geometric Generative Model.
\label{fig:GGM_framework}}
\end{figure}

\section{Geometric Generative Model}
\label{sec:GGM}

In this section, we propose a novel generative framework, which combines the discriminator and the generator together. The model decouples the two processes
\begin{enumerate}
\item Encoding/decoding process: This step maps the samples between the image space $\mathcal{X}$ and the latent (feature) space $\mathcal{Z}$ by using deep neural networks, the encoding map is denoted as $f_\theta:\mathcal{X}\to\mathcal{Z}$, the decoding map is $g_\xi: \mathcal{Z}\to\mathcal{X}$. This step achieves the dimension deduction.
\item Probability measure transformation process: this step transform a fixed distribution $\zeta\in \mathcal{P}(\mathcal{Z})$ to any given distribution $\mu\in \mathcal{P}(\mathcal{Z})$. The mapping is denoted as $T: \mathcal{Z}\to \mathcal{Z}$, $T_\#\zeta = \mu$. This step can either use conventional deep neural network or use explicit geometric/numerical methods.
\end{enumerate}
There are many existing methods to accomplish the encoding/decoding process, such as VAE model \cite{Kingma2013}, therefore we focus on the second step.

As shown in Fig.~\ref{fig:GGM_framework}, given an empirical distribution $\nu = \frac{1}{n} \sum_{i=1}^n \delta_{y_i}$ in the original ambient space $\mathcal{X}$, the support of $\nu$ is a sub-manifold $\Sigma \subset \mathcal{X}$. The encoding map $f_\theta:\Sigma\to \mathcal{Z}$ transform the support manifold $\Sigma$ to the latent (or feature) space $\mathcal{Z}$, $f_\theta$ pushes forward the empirical distribution to $\mu$ defined on latent space
\begin{equation}
    \mu = (f_\theta)_{\#} \nu = \frac{1}{n}\sum \delta_{z_i}.
\end{equation}
where $z_i=f_\theta(y_i)$.

Let $\zeta$ be a fixed measure on the latent space, we would like to find an optimal transportation map $T:\mathcal{Z}\to\mathcal{Z}$, such that $T_\# \zeta = \mu$.  This is equivalent to find the Brenier potential
\[
    u_h(z) = \max_{i=1}^n \{\langle z,z_i \rangle + h_i\}.
\]
Note that, $u_h$ can be easily represented by linear combinations and ReLus. The height parameter can be obtained by optimizing the energy Eqn.~\ref{eqn:Alexandrov_energy}
\[
    \frac{1}{n}\sum_{i=1}^k h_i - \int^h \sum_{i=1}^k w_i(\eta) d\eta_i.
\]
The optimal transporation map $T=\nabla u_h$. This can be carried out as a power diagram with weighted points $\{(z_i,\psi_i)\}$, where
\[
    \psi_i = \frac{|z_i|^2}{2} - h_i.
\]
The relation between the Kantorovich potential and the Brenier potential is
\[
    \varphi(x) = \frac{1}{2} |z|^2 - u_h(x).
\]
The Wasserstein distance can be explicitly given by
\[
    W_c(\zeta,\mu) = \int_{\mathcal{Z}} \varphi(z)d\zeta(z)
    + \frac{1}{n}\sum_{j=1}^n \psi_j.
\]
We use $g_\xi:\mathcal{Z}\to\mathcal{X}$ to denote the decoding map. Finally, the composition $g_\xi\circ T: \mathcal{Z}\to \mathcal{X}$ transforms $\zeta$ in the latent space to the original empirical distribution $\nu$ in the image space $\mathcal{X}$.

\section{Experiments}
\label{sec:experiments}
In order to demonstrate in principle the potential of our proposed method, we have designed and conducted the preliminary experiments.

\subsection{Comparison with WGAN}

\begin{figure}[t!hb]
 \centering
 \begin{tabular}{cc}
\includegraphics[width=0.485\textwidth]{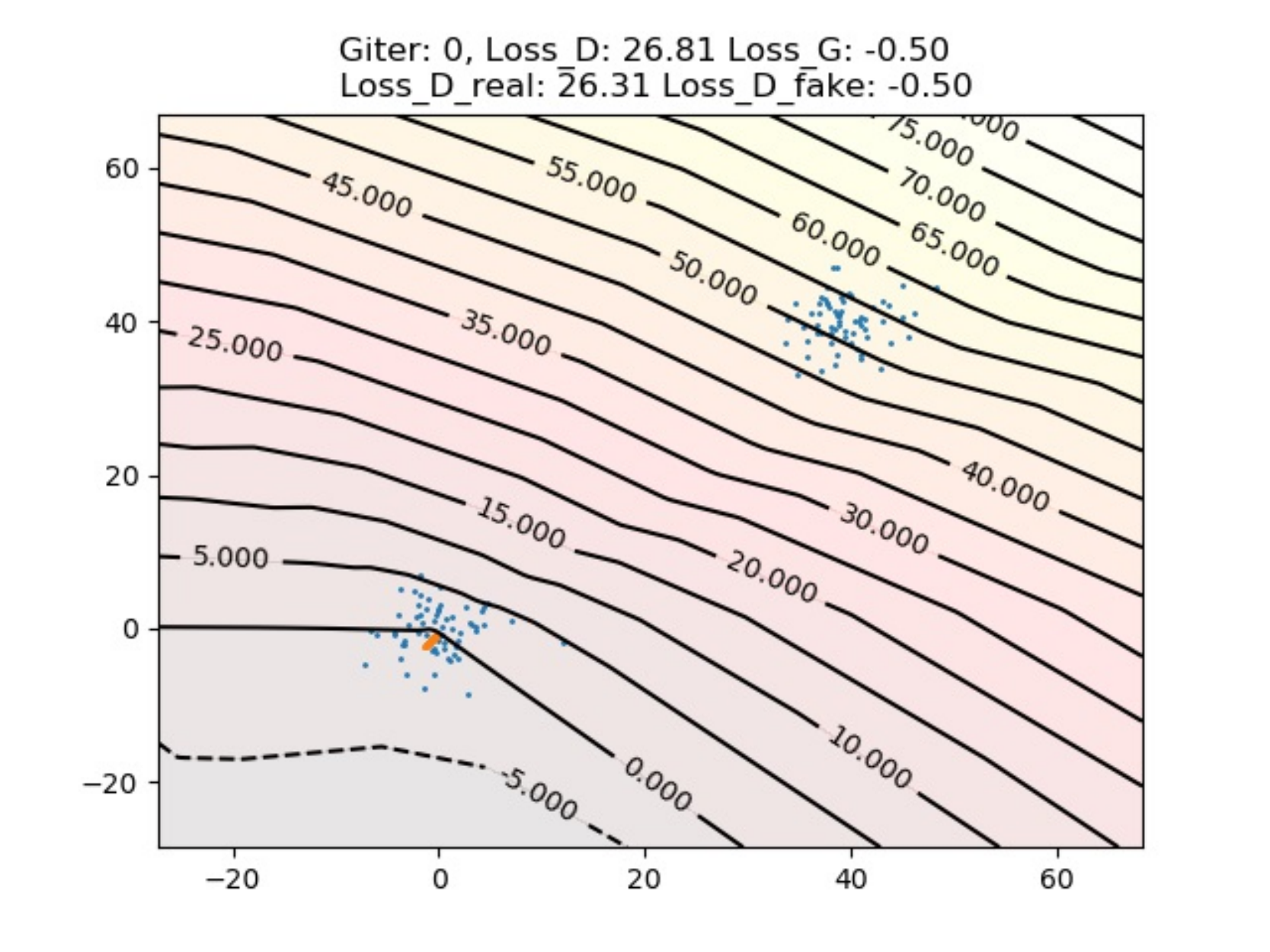}&
\includegraphics[width=0.485\textwidth]{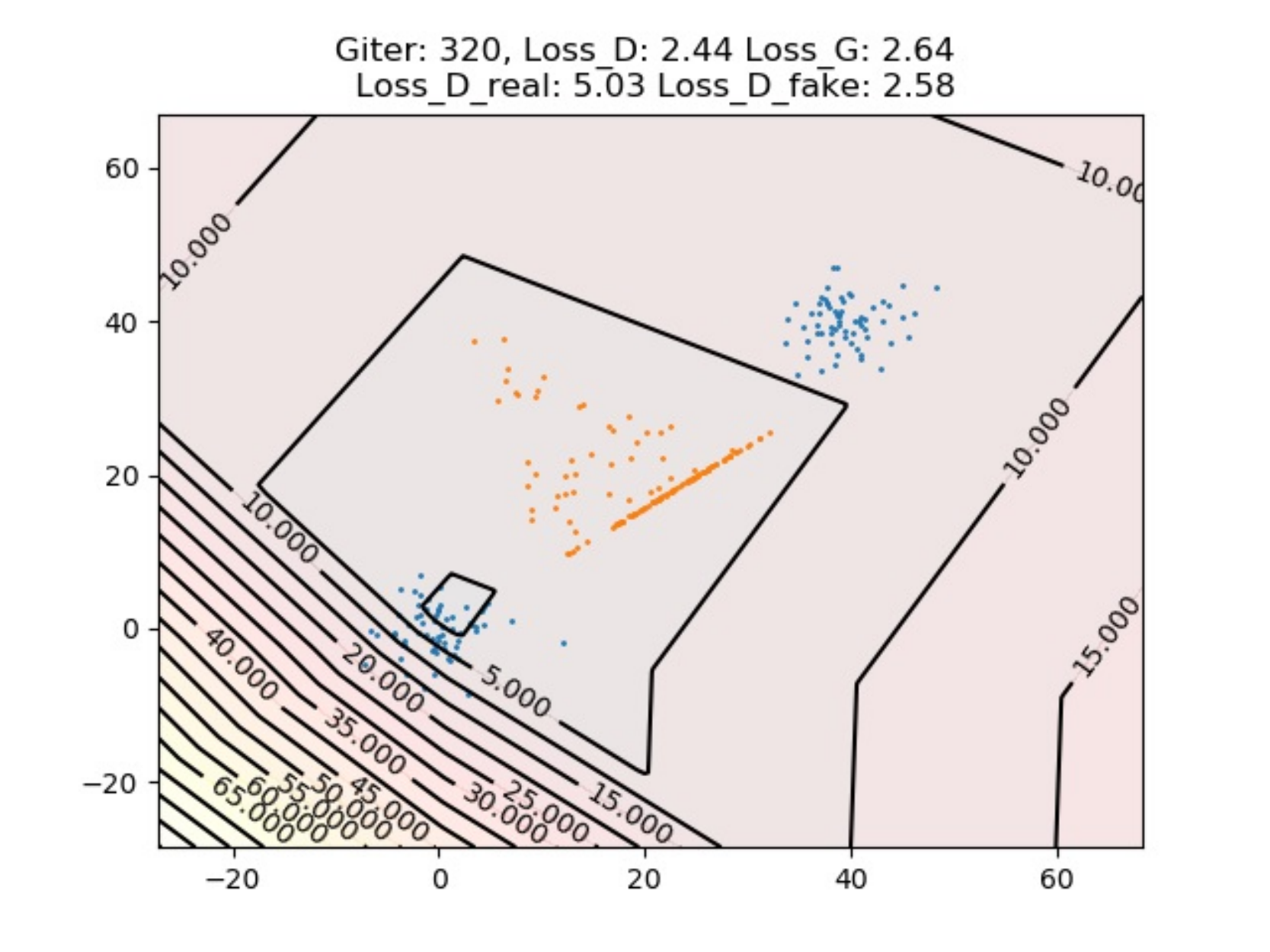}\\
(a) initial stage &(b) after $320$ iterations\\
\includegraphics[width=0.485\textwidth]{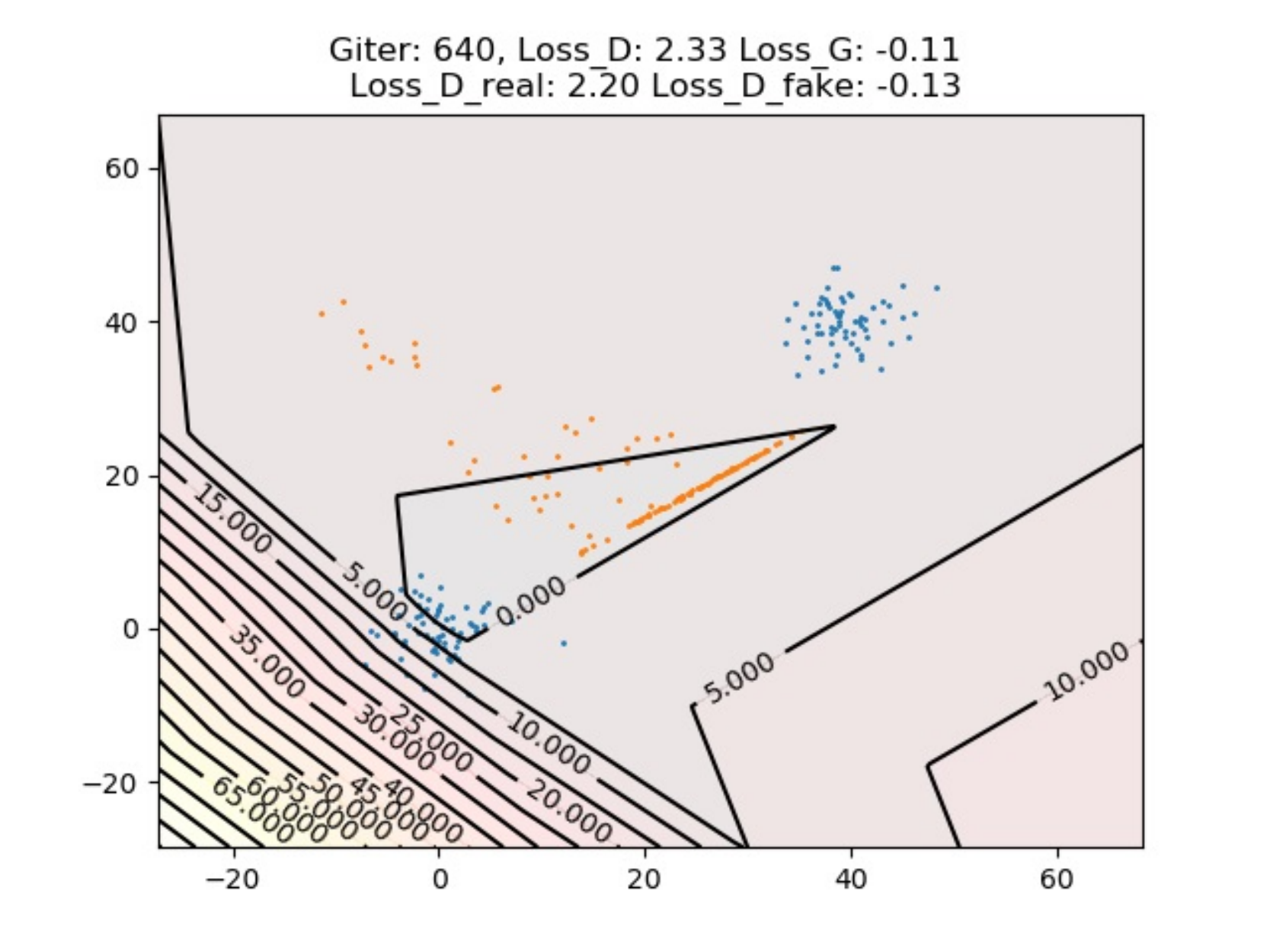}&
\includegraphics[width=0.485\textwidth]{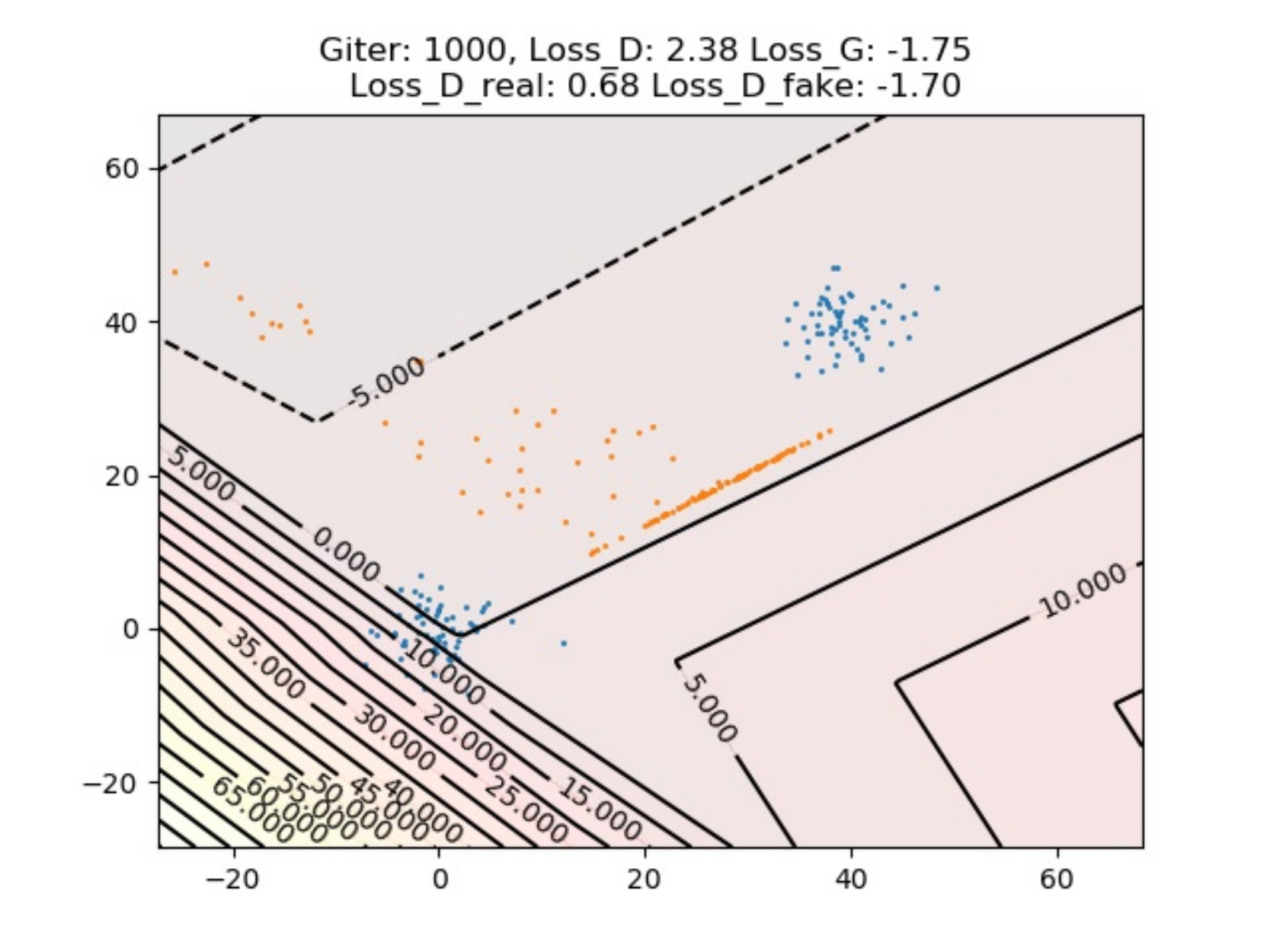}\\
(c) after $640$ iterations & (d) final stage, after $1000$ iterations
 \end{tabular}
 \caption{WGAN learns the Gaussian mixture distribution.
\label{fig:WGAN}}
\end{figure}
\begin{figure}[t!hb]
 \centering
 \begin{tabular}{cc}
\includegraphics[height=0.35\textwidth]{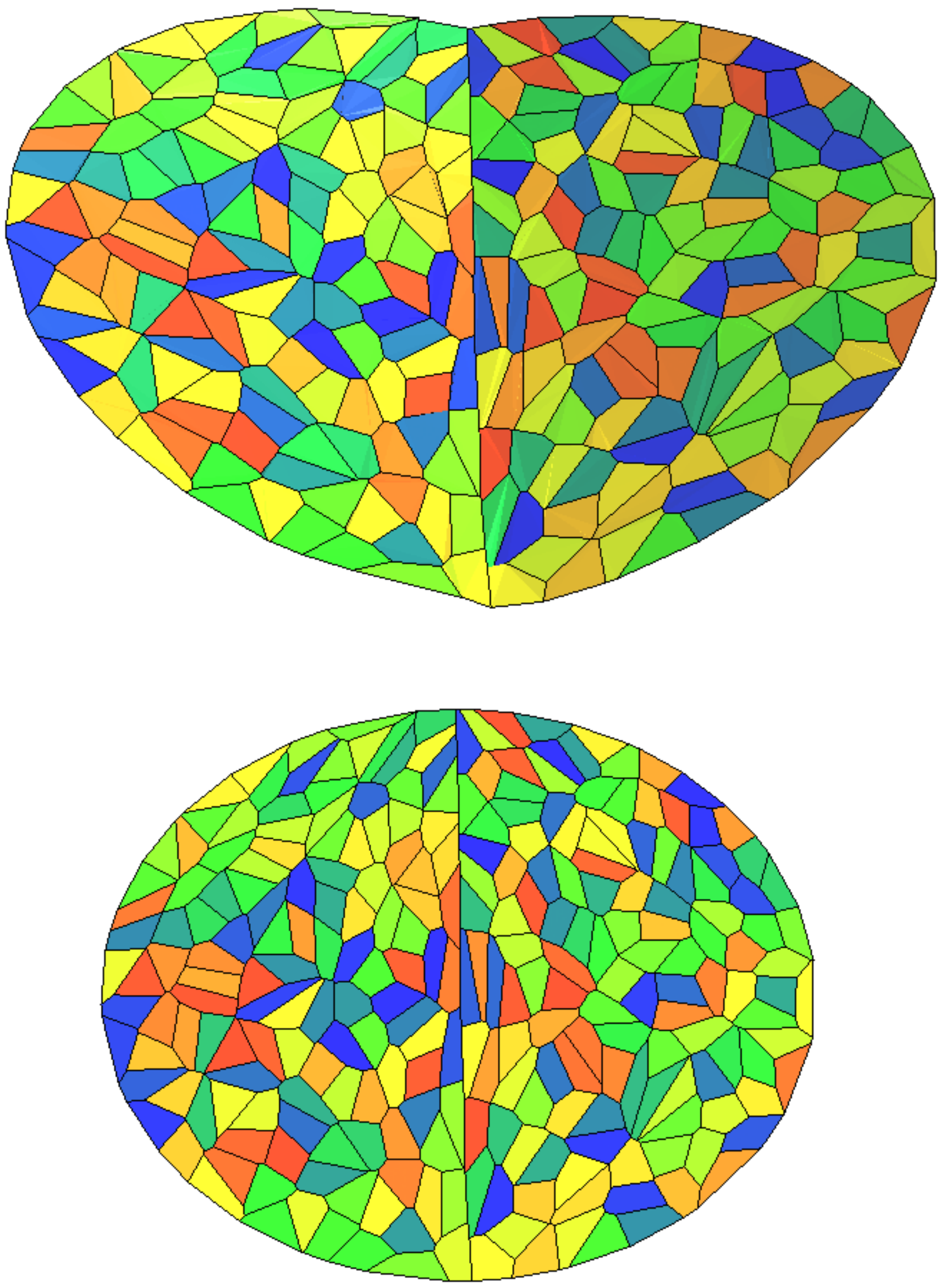}&
\includegraphics[height=0.35\textwidth]{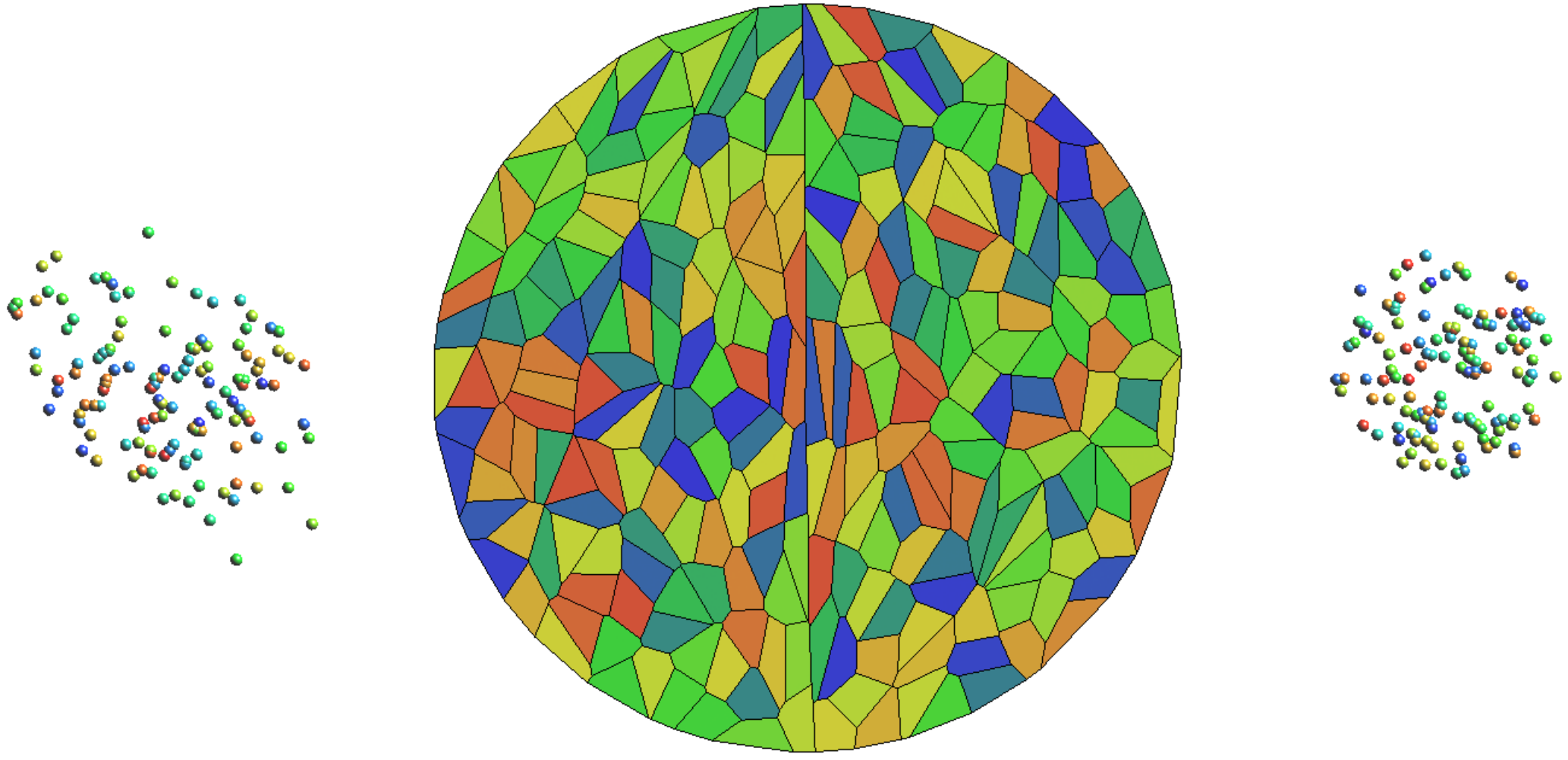}\\
(a) Brenier potential $u_h$ &(b) Optimal transportation map $T$: power diagram cell $W_i\mapsto y_i$\\
 \end{tabular}
 \caption{Geometric model learns the Gaussian mixture distribution .
\label{fig:Geom_GMD}}
\end{figure}

In the first experiment, we use Wasserstein Generative Adversarial Networks (WGANs) \cite{arjovsky2017wasserstein} to learn the mixed Gaussian distribution as shown in Fig.~\ref{fig:WGAN}.

\paragraph{Dataset}

The distribution of data $\nu$ is described by a point cloud on a $2d$ plane. We sample $128$ data points as real data from two Gaussian distributions, $\mathcal{N}(p_k,\sigma_k^2)$, $k=1,2$, where $p_1=(0,0)$ and $\sigma_1=3$, $p_2=(40,40)$ and $\sigma_2=3$. The latent space $\mathcal{Z}$is a square on the $2d$ plane $[1k,3k]\times[1k,3k]$, the input distribution $\zeta$ is the unform distribution on $\mathcal{Z}$. We use a generator to generate data from $\zeta$ to approximate the data distribution $\nu$. We generate $128$ samples in total.

\paragraph{Network Structure}
The structure of the discriminator is 2-layer ($2\times 10$ FC)-ReLU-($10\times 1$ FC) network, where FC denotes the fully connected layer. The number of inputs is $2$ and the number of outputs is $1$. The number of nodes of the hidden layer is $10$.

The structure of the generator is a 6-layer ($2\times 10$ FC)-ReLU-($10\times 10$ FC)-ReLU-($10\times 10$ FC)-ReLU-($10\times 10$ FC)-ReLU-($10\times 10$ FC)-ReLU-($10\times 2$ FC) network. The number of inputs is $2$ and the number of outputs is $2$. The number of nodes of all the hidden layer is $10$.

\paragraph{Parameter Setting}
For WGAN, we clip all the weights to $[-0.5, 0.5]$. We use the RMSprop \cite{Hinton_2012} as the optimizer for both discriminator and generator. The leaning rate of both the discriminator and generator are set to $1e-3$.

\paragraph{Deep learning framework and hardware}
We use the PyTorch \cite{pytorch} as our deep learning tool. Since the toy dataset is small, we do experiments on CPU. We perform experiments on a cluster with $48$ cores and $193GB$ RAM. However, for this toy data, the running code only consumes $1$ core with less than $500MB$ RAM, which means that it can run on a personal computer.

\paragraph{Results analysis}
In Fig.~\ref{fig:WGAN}, the blue points represent the real data distribution and the orange points represent the generated distribution. The left frame shows the initial stage, the right frame illustrates the stage after 1000 iterations. It seems that WGAN cannot capture the Gaussian mixture distribution. Generated data tend to lie in the middle of the two Gaussians. One reason is the well known mode collapse problem in GAN, meaning that if the data distribution has multiple clusters or data is distributed in multiple isolated manifolds, then the generator is hard to learn multiple modes well. Although there are a couple of methods proposed to deal with this problem
\cite{Gurumurthy_CVPR_2017,Hoang_arXiv_2017}, these methods require the number of clusters, which is still a open problem in the machine learning community.

\paragraph{Geometric OMT} Figure~\ref{fig:Geom_GMD} shows the geometric method to solve the same problem. The left frame shows the Brenier potential $u_h$, namely the upper envelope, which projects to the power diagram $\mathcal{V}$ on a unit disk $\mathbb{D}\subset \mathcal{Z}$, $\mathcal{V}=\bigcup_k W_i(h)$. The right frame shows the discrete optimal transportation map $T:\mathbb{D}\to \{y_i\}$, which maps each cell $W_i(h)\cap \mathbb{D}$ to a sample $y_i$, the cell $W_i(h)$ and the sample $y_i$ have the same color. All the cells have the same area, this demonstrates that $T$ pushes the uniform distribution $\zeta$ to the exact empirical distribution $T_\#\zeta = 1/n \sum_i \delta_{y_i}$.

The samples $\{y_i\}$ are generated according to the same Gauss mixture distribution, therefore there are two clusters. This doesn't cause any difficulty for the geometric method. In the left frame, we can see the upper envelope has a sharp ridge, the gradients point to the two clusters. Hence, the geometric method outperforms the WGAN model in the current experiment.

\subsection{Geometric Method}

\begin{figure}[t!hb]
 \centering
 \begin{tabular}{cc}
\includegraphics[width=0.45\textwidth]{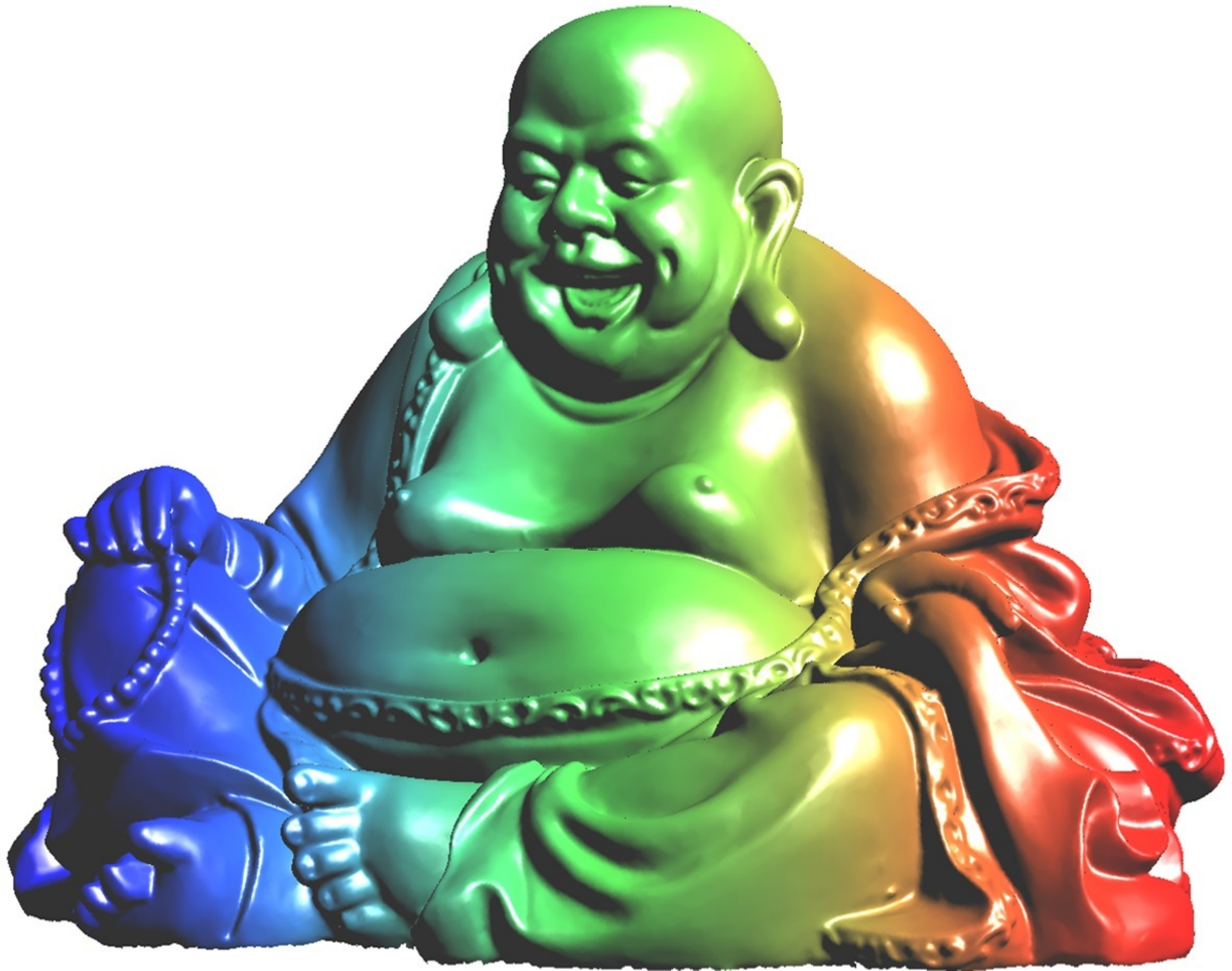}&
\includegraphics[width=0.45\textwidth]{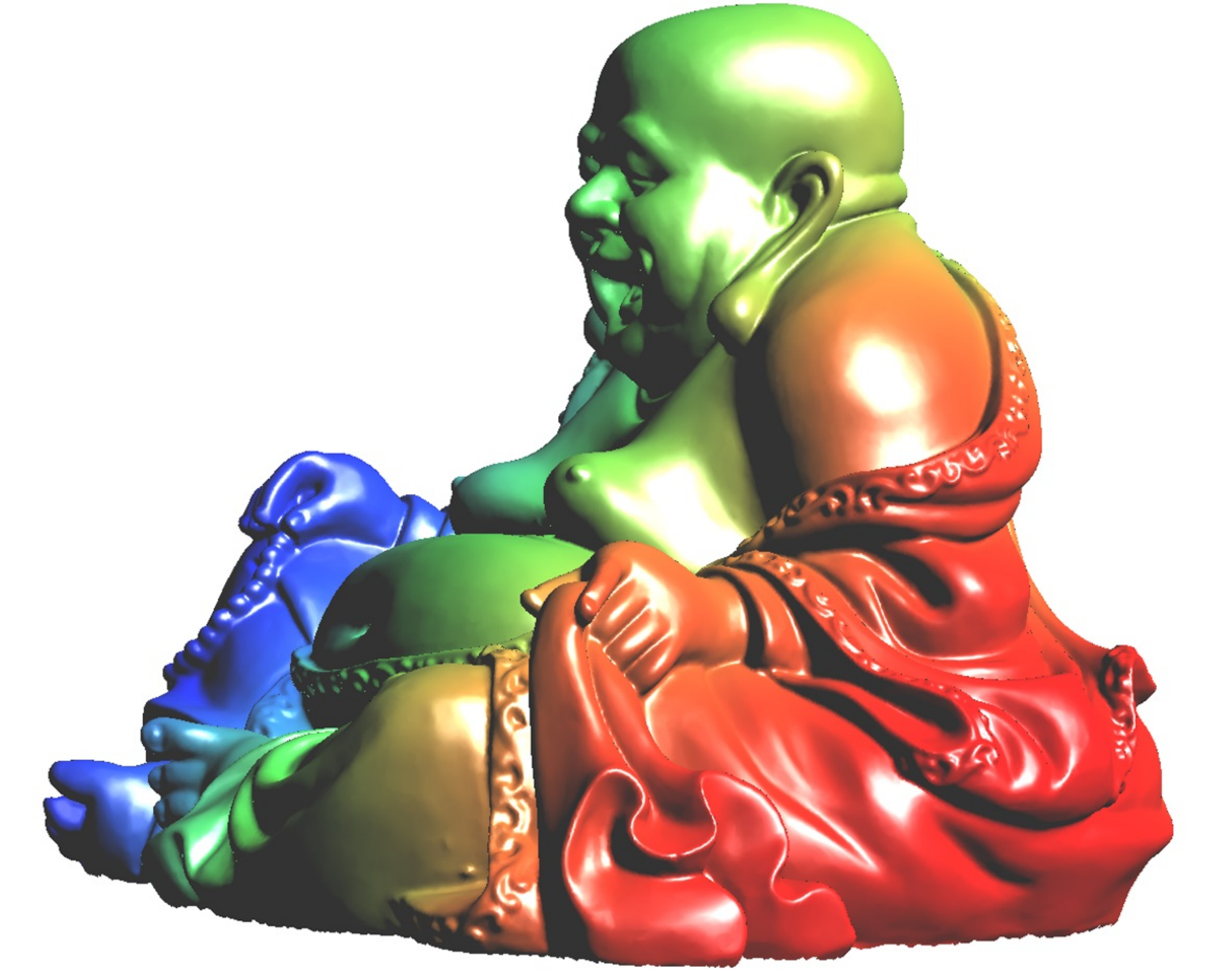}\\
(a) Supporting manifold $\Sigma$ &(b) Supporting manifold $\Sigma$\\
\includegraphics[width=0.45\textwidth]{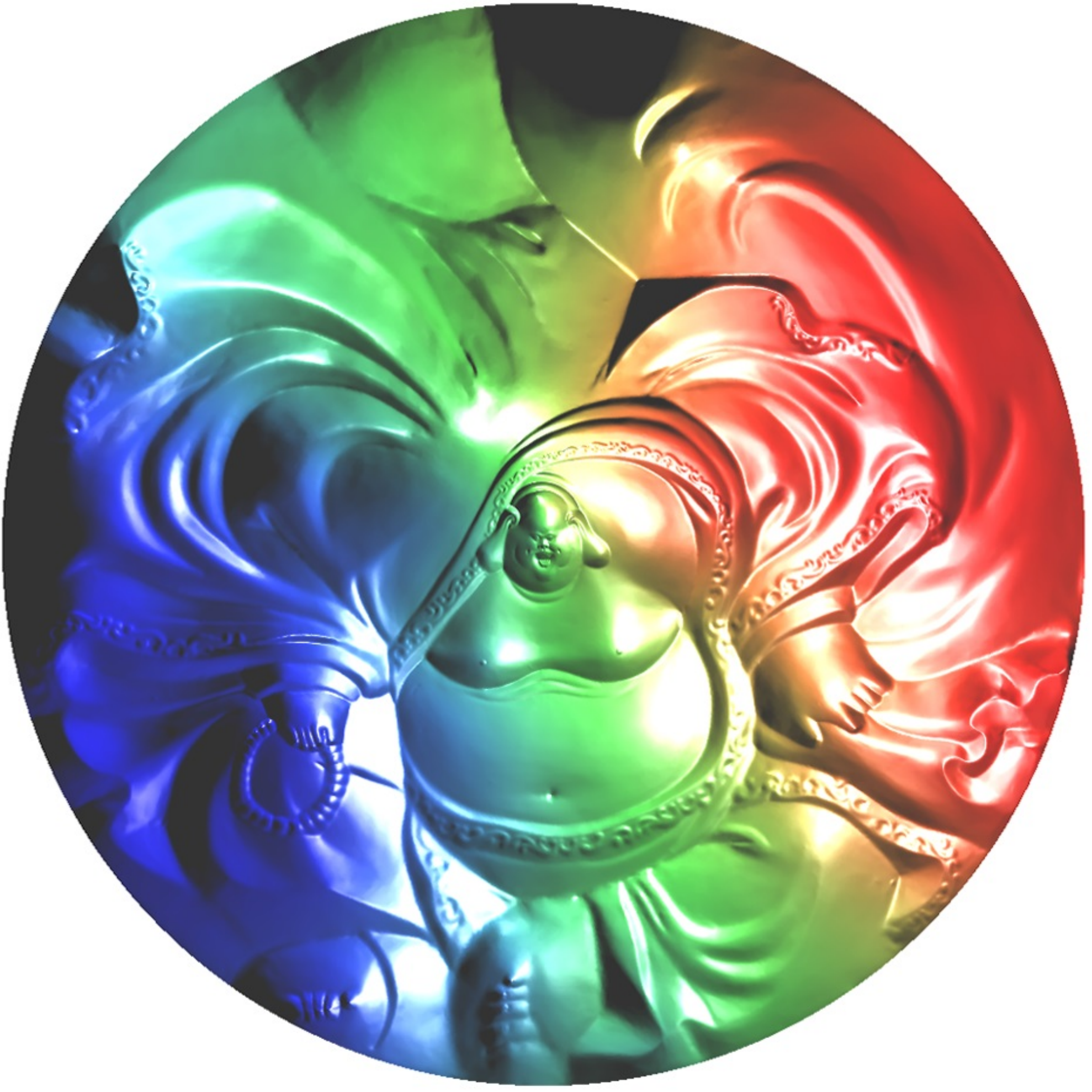}&
\includegraphics[width=0.45\textwidth]{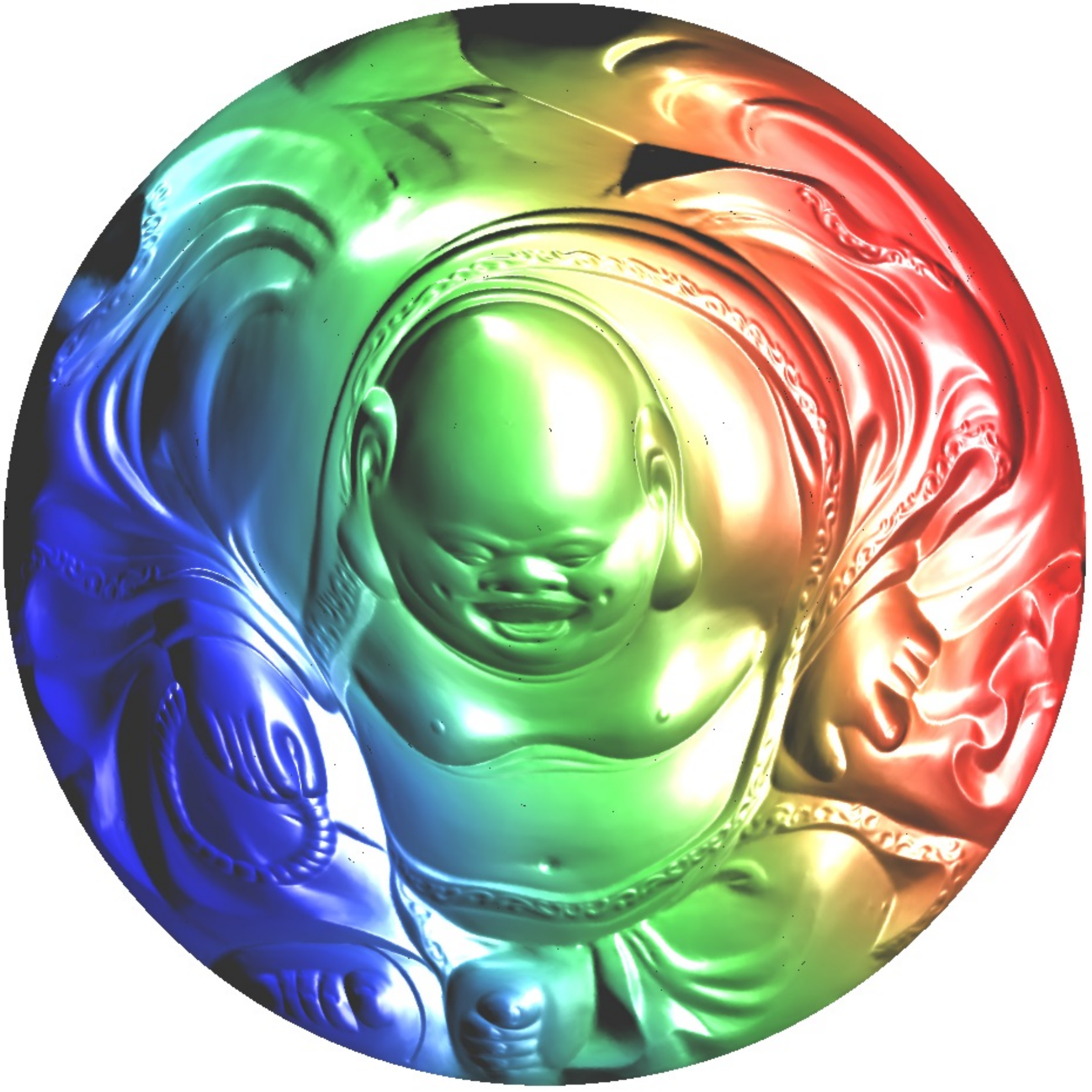}\\
(c) Image of encoding map $f_\theta(\Sigma)$ &(d) Image of encoding map composed with\\
&  optimal transportation map $(T^{-1} \circ f_\theta)(\Sigma)$\\
 \end{tabular}
 \caption{Illustration of geometric generative model.
\label{fig:GGM_1}}
\end{figure}

\begin{figure}[t!hb]
 \centering
 \begin{tabular}{cc}
\includegraphics[width=0.4\textwidth]{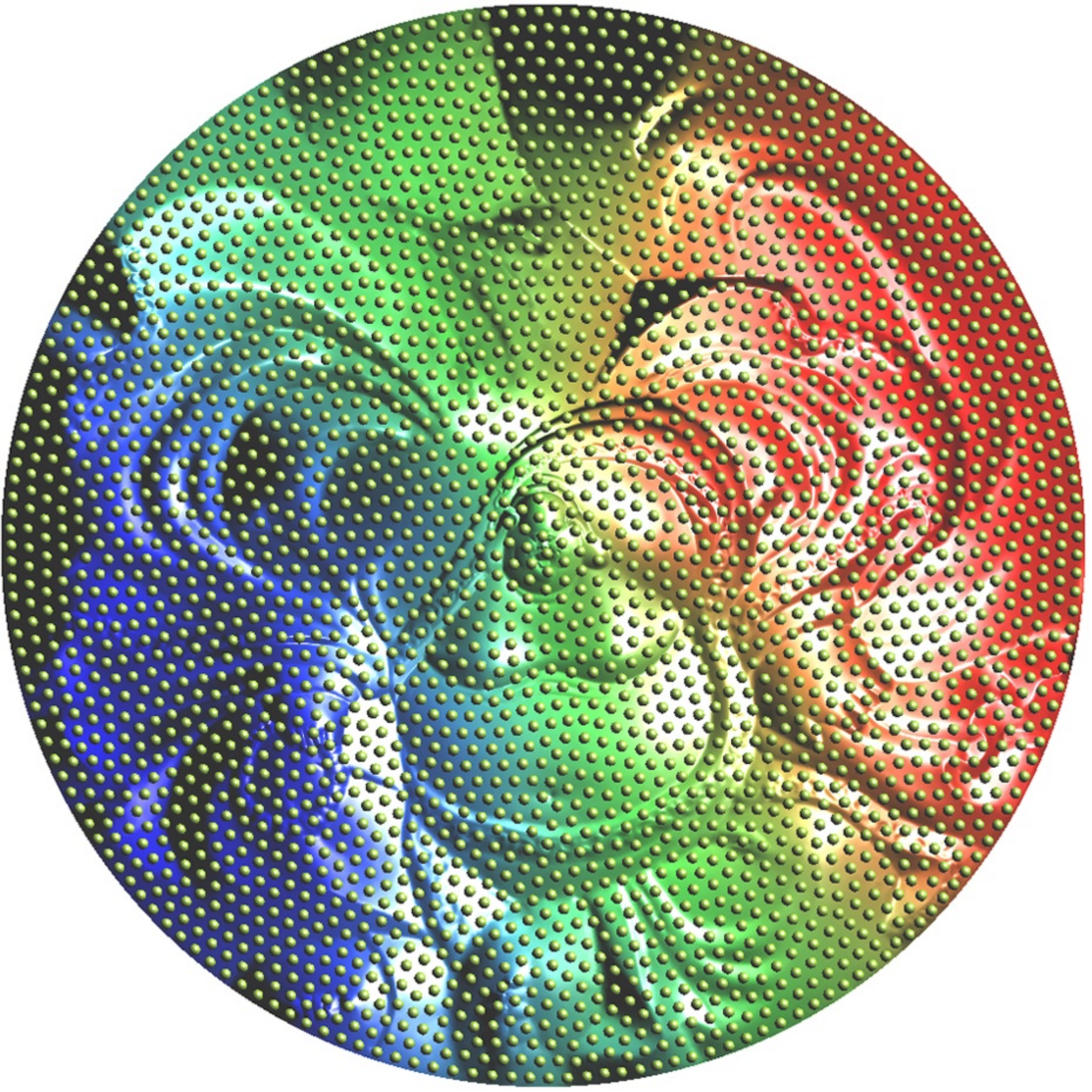}&
\includegraphics[width=0.475\textwidth]{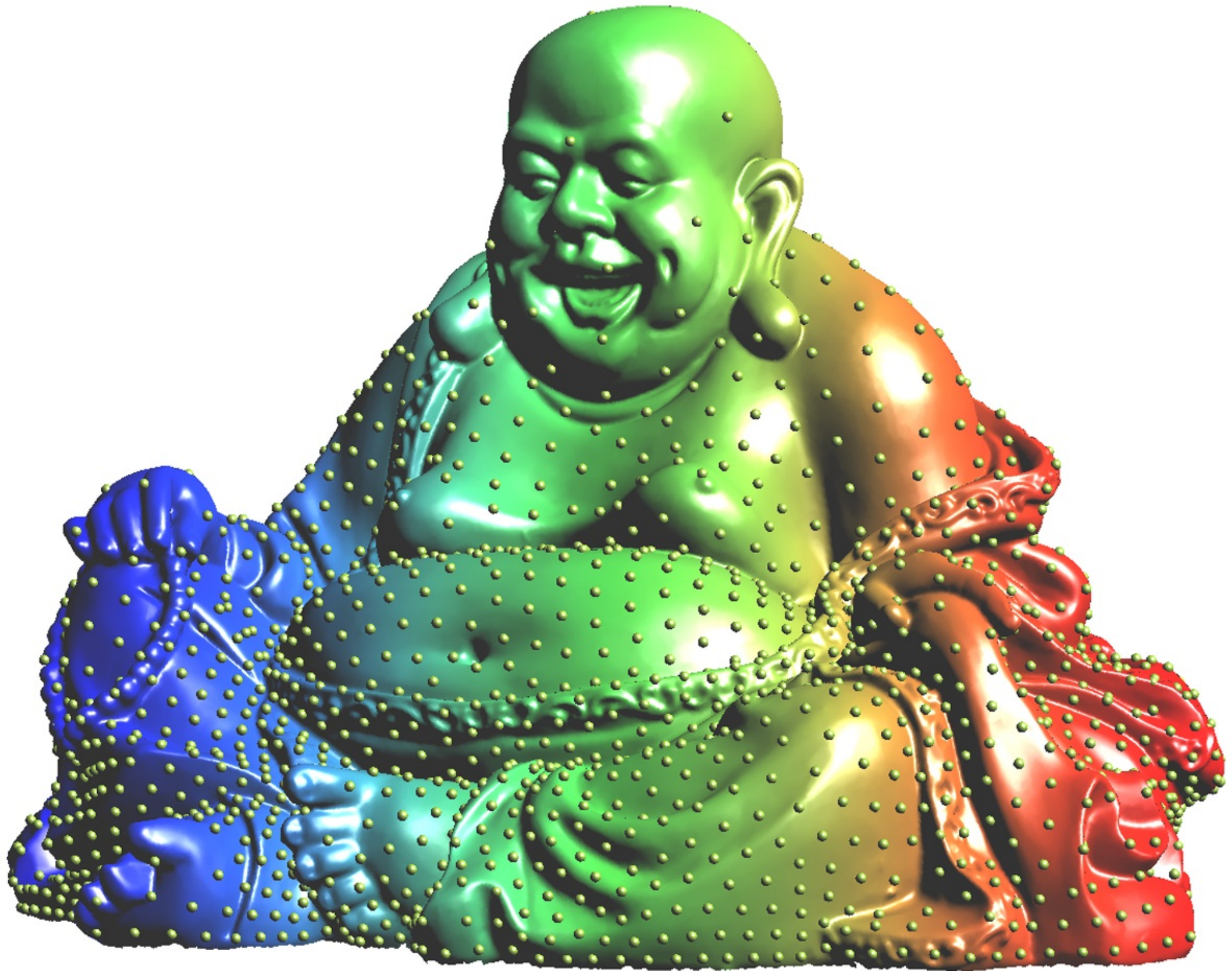}\\
(a) sampling according to the & (b) non-uniform sampling according to the\\
uniform distribution $\zeta$ on $\mathcal{Z}$  & distribution $(f^{-1}_\theta)_\#\zeta$ on $\Sigma$\\
\includegraphics[width=0.4\textwidth]{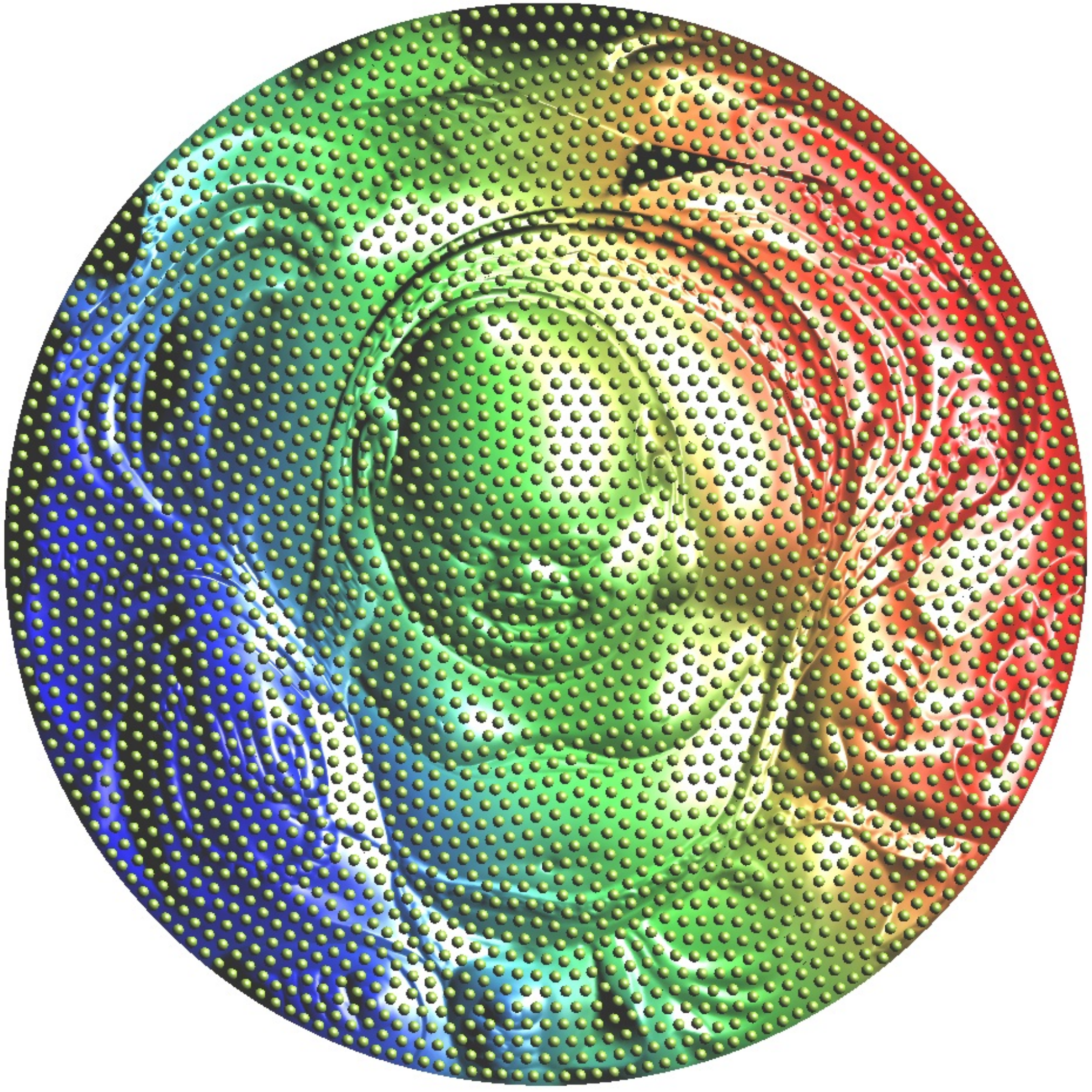}&
\includegraphics[width=0.475\textwidth]{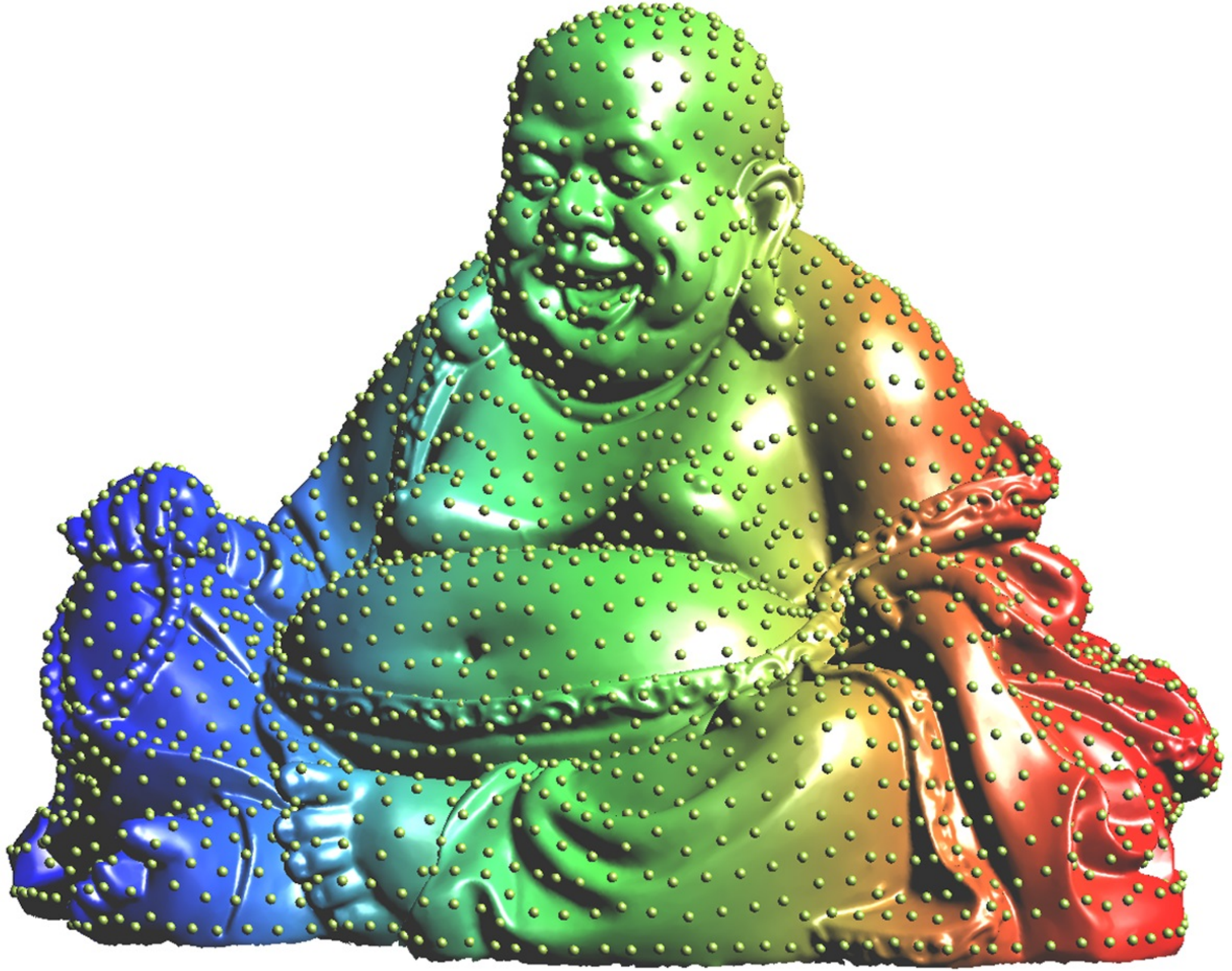}\\
(c) sampling according to the& (d) uniform sampling according to \\
uniform distribution $\zeta$ on $\mathcal{Z}$  & $(f^{-1}_{\theta}\circ T)_\#\zeta$ on $\Sigma$\\

 \end{tabular}
 \caption{Illustration of geometric generative model.
\label{fig:GGM_2}}
\end{figure}

In this experiment, we use pure geometric method to generate uniform distribution on a surface $\Sigma$ with complicated geometry. As shown in Fig.~\ref{fig:GGM_1}, the image space $\mathcal{X}$ is the $3$ dimensional Euclidean space $\mathbb{R}^3$. The real data samples are distributed on a submanifold $\Sigma$, which is represented as a surface, as illustrated in (a) and (b). The encoding mapping $f_\theta:\Sigma\to\mathcal{Z}$ maps the supporting manifold to the latent space, which is a planar disk. The encoding map $f_\theta$ can be computed using discrete surface Ricci flow method \cite{Gu_JDG_2017}. We color-encode the normals to the surface, and push forward the color function from $\Sigma$ to the latent space $f_\theta(\Sigma)$, therefore users can directly visualize the correspondence between $\Sigma$ and its image in $\mathcal{Z}$ as shown in (c). Then we construct the optimal mass transportation map $T: \mathcal{Z}\to \mathcal{Z}$, the image is shown in (d).

In Fig.~\ref{fig:GGM_2}, we demonstrate the generated distributions. In (a), we generate samples $\{z_1,\dots,z_k\}$ on the latent space $f_\theta(\Sigma)$ according to the uniform distribution $\zeta$, the samples are pulled back to the surface $\Sigma$ as $\{f_\theta^{-1}(z_1),\dots, f_\theta^{-1}(z_k)\}$ as shown in (b), which illustrate the distribution $(f_\theta^{-1})_\#\zeta$. It is obvious that the distribution generated this way is highly non-uniform on the surface. In frame (c), we uniformly generate samples on $(T^{-1}\circ f_\theta)(\Sigma)$, and map them back to the surface $\Sigma$ as shown in (d). This demonstrates the generated distribution $(f_\theta^{-1}\circ T)_\# \zeta$ on $\Sigma$, which is uniform as desired.

\section{Discussion and Conclusion}
\label{sec:discussion_conclusion}

In this work, we bridge convex geometry with optimal transportation, then use optimal transportation to analyze generative models. The basic view is that the discriminator computes the Wasserstein distance or equivalently the Kantorovich potential $\varphi_\xi$; the generator calculates the transportation map $g_\theta$. By selecting the transportation cost, such as $L^p$, $p>1$ distance, $\varphi_\xi$ and $g_\theta$ are related by a closed form, hence it is sufficient to train one of them.

For general transportation cost $c(x,y)$, the explicit relation between $\varphi_\xi$ and $g_\theta$ may not exist, it seems that both training processes are necessary. In the following, we argue that it is still redundant. For a given cost function $c$, the optimal decoding map is $g_1$; by using $L^2$ cost function, the solution is $g_0$. Both $g_0$ and $g_1$ induces the same measure,
\[
    (g_0)_\#\zeta = (g_1)_\# \zeta = \nu,
\]
By Brenier polar factorization theorem~\ref{thm:Brenier_factorization}, $g_1=g_0\circ s$, where $s:\mathcal{Z}\to\mathcal{Z}$, preserves the measure $\zeta$, $s_\# \zeta = \zeta$. All such of mappings form a infinite dimensional group, comparing to $g_0$, the complexity of $s$ increases the difficulty of finding $g_1$. But both $g_0$ and $g_1$ generate the same distribution, there is no difference in terms of the performance of the whole system. It is much more efficient to use $g_0$ without the double training processes.

For high dimensional setting, rigorous computational geometric method to compute the optimal transportation map is intractable, due to the maintenance of the complex geometric data structures. Nevertheless, there exist different algorithms to handle high dimensional situation, such as socialistic method, sliced optimal transportation method, hierarchical optimal transportation method.

In the future, we will explore along this direction, and implement the proposed model in a large scale.

\section*{Acknowledgement}

We thank the inspiring discussions with Dr. Dimitris Samaras from Stony Brook University, Dr. Shoucheng Zhang from Stanford University, Dr. Limin Chen from Ecole Centrale de Lyon and so on. The WGAN experiment is conducted by Mr. Huidong Liu from Stony Brook University. The project is partially supported
by NSFC  No. 61772105, 61772379, 61720106005, NSF DMS-1418255, AFOSR FA9550-14-1-0193 and the Fundamental Research Funds for the Central Universities No. 2015KJJCB23.

\section*{Appendix}

\subsection{Commutative Diagram}
The relations among geometric/functional objects are summarized in the following diagram:
\[
\begin{CD}
\mathcal{A} @>\text{Legendre dual}>> \mathcal{C}\\
@AA \text{integrate} A @AA A\\
u_h @>\text{Legendre dual}>> u_h^*\\
@VV \text{graph} V @VV \text{graph} V\\
\text{Env}(\{\pi_i\}) @>\text{Poincare dual}>> \text{Conv}(\{\pi_i^*\})\\
@VV \text{proj} V @VV \text{proj} V\\
\mathcal{V}(\psi) @>\text{Poincare dual}>> \mathcal{T}(\psi)
\end{CD}
\]
where each two adjacent layers are commutable. These relations can be observed from Fig.~\ref{fig:discrete_VP_framework} as well.

\subsection{Symbol list}

The following is the symbol list
\begin{table}[h!]
  \centering
  \caption{Symbol list.}
  \label{tab:table1}
  \begin{tabular}{lll}
  \hline\hline
  $\mathcal{X}$ &ambient space, image space & \\
  $\Sigma$ & support manifold for some distribution&\\
  $\mathcal{Z}$ & latent space, feature space &\\
  $\zeta$ & a fixed probability measure on $\mathcal{Z}$&\\
  $g_\theta$ & generating map& $g_\theta: \mathcal{Z}\to\mathcal{X}$\\
  $\varphi_\xi$ & Kantorovich potential &\\
  $c$ & distance between two points & $c(x,y)=|x-y|^p$, $p\ge 1$ \\
  $W_c$& Wasserstein distance & $W_c(\mu,\nu)$\\
  \hline
  $X$ & source space&\\
  $Y$ & target space&\\
  $\mu$ & source probability measure&\\
  $\nu$ & target probability measure&\\
  $\Omega$ & source domain & $\Omega\subset X$\\
  $y_i$ & the i-th sample in target& $\{y_1,\dots,y_k\}\in Y$ \\
  $\varphi$& Kantorovich potential& $\phi^c=\psi$, $\psi^c = \phi$\\
  $\psi$& power weight & $\psi=(\psi_1,\dots,\psi_k)$\\
  $h$   & plane heights& $h=(h_1,\dots,h_k)$\\
  $\pi_i$ & hyper-plane & $\pi_i^h(x)=\langle y_i,x\rangle + h_i$\\
  $\pi_i^*$ & dual point of $\pi_i$ & $\pi_i^*=(y_i,-h)$\\
  $\text{pow}$ & power distance & $\text{pow}(x,y_i)=c(x,y_i)-\psi_i$\\
  $W_i$ & power voronoi cell & $W_i(\psi) = \{x\in X| \text{pow}(x,y_i)\le \text{pow}(x,y_j)\}$\\
  $w_i$ & the volume of $W_i$ & $w_i(h) = \mu(W_i(h)\cap \Omega)$\\
  $u$   & Brenier potential & $u_h(x)= \max_i \{ \langle x, y_i \rangle + h_i \}$\\
  $\mathcal{A}$ & Alexandrov potential & $\mathcal{A}(h)=\int^h \sum_i w_i dh_i$ \\
  $T$ & transportation map & $T=\nabla u_h$\\
  $\mathcal{C}$ & transportation cost & $\mathcal{C}(T)=\int_X c(x,T(x)) d\mu$ \\
  $\text{Env}$ & upper envelope of planes & $\text{Env}(\{\pi_i\})$ graph of $u_h$\\
  $\text{Conv}$ & convex hull of points & $\text{Conv}(\{\pi_i^*\})$ graph of $u_h^*$\\
  $\mathcal{V}$ & power diagram & $\mathcal{V}(\psi):X = \bigcup_i W_i(\psi)$\\
  $\mathcal{T}$ & weighted Delaunay triangulation\\
  \hline\hline
  \end{tabular}
\end{table}


\end{document}